\documentclass[10pt]{article} 
\usepackage[nocompress]{cite}
\usepackage{array}
\usepackage{mdwmath}
\usepackage{mdwtab}
\usepackage{eqparbox}
\usepackage[hyphens,spaces,obeyspaces]{url}
\usepackage{amsmath}
\usepackage{float, xcolor}
\usepackage{graphicx}
\usepackage{multirow}
\usepackage{subfig}
\usepackage{booktabs}
\usepackage{amsmath,graphicx}
\usepackage{amsxtra}
\usepackage{sectsty}
\usepackage{threeparttable}
\usepackage[breaklinks=true]{hyperref}
\usepackage{bm}
\usepackage{longtable}
\usepackage{enumitem}
\usepackage{ntheorem}
\usepackage{booktabs}
\usepackage{CJKutf8}
\usepackage{booktabs}
\usepackage{multirow}
\usepackage{subfig}
\usepackage{authblk}

\begin{document}
\title{Natural Language Processing in Electronic Health Records in Relation to Healthcare Decision-making: A Systematic Review}

\author[1*]{Elias Hossain } 
\author[2]{Rajib Rana}
\author[3,4,5]{Niall Higgins}
\author[6]{Jeffrey Soar}
\author[6]{Prabal Datta Barua}
\author[7]{Anthony R. Pisani, Ph.D}
\author[4]{Kathryn Turner}

\affil[1]{School of Electrical and Computer Engineering, North South University, Dhaka – 1229, Bangladesh}

\affil[2]{School of Mathematics, Physics and Computing, University of Southern Queensland, Springfield Central QLD 4300, Australia}

\affil[3]{School of Management and Enterprise, University of Southern Queensland, Darling Heights QLD 4350, Australia}

\affil[4]{School of Nursing, Queensland University of Technology, Kelvin Grove, Brisbane, QLD 4000, Australia}

\affil[5]{Metro North Mental Health, Herston QLD 4029, Australia} 

\affil[6]{School of Business, University of Southern Queensland,  Springfield Central QLD 4300, Australia}

\affil[7]{Center for the Study and Prevention of Suicide, University of Rochester, Rochester, NY, United States}

\affil[*]{Corresponding author: elias.hossain191@gmail.com}






\vspace*{.01 in}
\maketitle
\vspace{.12 in}
\vspace{5mm}

\section*{abstract}
\vspace{5mm}
\textbf{Background:} Natural Language Processing (NLP) is widely used to extract clinical insights from Electronic Health Records (EHRs). However, the lack of annotated data, automated tools, and other challenges hinder the full utilisation of NLP for EHRs. Various Machine Learning (ML), Deep Learning (DL) and NLP techniques are studied and compared to understand the limitations and opportunities in this space comprehensively.

\textbf{Methodology:} 
After screening 261 articles from 11 databases, we included 127 papers for full-text review covering seven categories of articles: 1) medical note classification, 2) clinical entity recognition, 3) text summarisation, 4) deep learning (DL) and transfer learning architecture, 5) information extraction, 6) Medical language translation and 7) other NLP applications. This study follows the Preferred Reporting Items for Systematic Reviews and Meta-Analyses (PRISMA) guidelines.

\textbf{Result and Discussion:} EHR was the most commonly used data type among the selected articles, and the datasets were primarily unstructured. Various ML and DL methods were used, with prediction or classification being the most common application of ML or DL. The most common use cases were: the International Classification of Diseases, Ninth Revision (ICD-9) classification, clinical note analysis, and named entity recognition (NER) for clinical descriptions and research on psychiatric disorders.

\textbf{Conclusion:} 
We find that the adopted ML models were not adequately assessed. In addition, the data imbalance problem is quite important, yet we must find techniques to address this underlining problem. Future studies should address key limitations in studies, primarily identifying Lupus Nephritis, Suicide Attempts, perinatal self-harmed and ICD-9 classification.

\section*{keywords} 
Machine learning; electronic health records; medical natural language processing; artificial intelligence in medicine ; automated tools ; state-of-the-art deep learning

\vspace{.12 in}


\section{introduction}

Electronic Health Records (EHRs), which are automated compilations of health care activities and assessments, are increasingly prevalent and essential for healthcare provision, administration, and research \cite{cowie2017electronic}. The data found in EHRs can be both structured and unstructured  \cite{consultant2015unstructured}. Structured EHR data comprises heterogeneous sources in fixed numerical or categorical areas, such as diagnoses, prescriptions, and laboratory values. On the other hand, Produced by healthcare personnel, clinical documentation or note and discharge summaries represent instances of unstructured data. Clinical documentation or notes are input as free text into EHRs, offering a complete picture of the patient's condition.

The adoption of EHRs has increased rapidly around the world. In the United States, it has increased dramatically from 10\% to nearly 96\% in just 10 years (2008-2017). In China, this increase is slightly more than 85\% \cite{Liang2021Adoption}. A similar trend has been observed in General Practices, large hospitals, and health services in Australia \cite{Hodgkins2020Australian,Cairns2021Building}. 

With the increased adoption of EHRs, the volume of information can now be considered "Big Data," extending to the modification and application of substantial data accumulated in EHRs. However, the capacity of human cognition to study, comprehend, and interpret data is constrained; therefore, there is a need to contrive computer-based tools that can organize, evaluate, and recognize patterns within these data. The subsequent step is to convert all of these extensive healthcare data into knowledge by implementing data mining and natural language processing methods as essential components in data analytics on EHRs big data to aid the development of an EHR ecosystem.

Most innovations for this growing number of unstructured free texts in the medical domain are based on novel Machine Learning (ML), and Deep Learning (DL) techniques \cite{naseem2021comparative} \cite{bhosale2022application}. Numerous healthcare and medical applications\cite{bhosale2022iot} include the detection of cardiovascular risk factors and heart conditions \cite{beam2018big}, the diagnosis \cite{bhosale2022deep} and prognosis of oral diseases \cite{leite2020radiomics}, and the detection of cancer tumours from radiology images \cite{esteva2019guide}, have made extensive use of machine learning. Recently, the concept of autoML, one of the types of ML integrated tool \cite{feurer2015efficient}, has been presented as a way to expand the applications of ML algorithms and simplify the implementation of those algorithms in a range of industries, including in healthcare \cite{waring2020automated}. Although AutoML is still an emerging technology, it has already been applied in bioinformatics, translational medicine, diabetes diagnosis, Alzheimer diagnosis, electronic health record (EHR) analysis, and imaging for medical purposes \cite{borkowski2019google}. However, it has not been extensively investigated for how it might be used to process clinical notes, a significant component of EHR. In addition, patient data protection is a significant concern when employing ML-enabled automated systems; yet, no research has identified patient data protection difficulties or comprehensively explored the strategies that may be implemented to assure medical data privacy.

Aside from these, several solutions have been developed for EHRs to handle clinical tasks; however, there remain challenges for health information research because of the unique language and clinical idioms used by clinicians \cite{choudhary2022nlp,Ulrich2021The}. Natural Language Processing (NLP), a subfield of Artificial Intelligence (AI) techniques (such as entity recognition), has been used for clinical text mining \cite{afzal2018natural,Galatzan2021Testing}, which is a notably clinical note analysis. Theoretically, these techniques are in their conception stage, and it will take some time for them to be able to select an accurate and precise model for real-world applications. This leads to the most significant problem in the field of NLP: the processing of medical text data and decision-making utilising computer technologies. There is a need for novel ways to classify NLP to facilitate its effective use in contemporary healthcare. This project's first and foremost objective is to solve the identified shortcomings in EHRs-NLP applications for healthcare and find effective methods for analyzing EHRs, which will have a positive influence on the research community.

This study provides a thorough review of the numerous healthcare uses of NLP. The objectives of our review are as follows. First, we aim to review the NLP technique in EHRs with a specific focus on different state-of-the-art models. Second, we explain the DL and ML paradigms used to analyse EHRs, mainly clinical free text. Third, we identify core challenges in categorising clinical notes. Last, we examine how researchers have implemented their models for managing clinical notes in the healthcare industry.

We present the difference between our review and existing ones in Table~\ref{tab:comparison_table}. It is noted that a high proportion of review articles used NLP and EHRs. Still, few seem to have adhered to the PRISMA structure, an evidence-based minimal set of reporting elements for comprehensive meta-analyses and reviews. Recent reviews have covered DL and ML-based strategies, but reviews published before 2020 have not emphasised the use of state-of-the-art models or explained many potential challenges in clinical NLP. Similarly, information regarding model validation or evaluation matrix was missing. Finally, none of the existing reviews discussed clinical tools or settings, let alone advanced NLP methods such as the transformer model.

\vspace{5mm}
\begin{table}[h]
\caption{Comparison of our paper with that of the existing articles}
\label{tab:comparison_table}
\resizebox{\columnwidth}{!}{%
\begin{tabular}{|c|c|c|c|c|c|c|c|c|c|c|c|}
\hline
Year &
  Authors &
  \begin{tabular}[c]{@{}c@{}}PRISMA\\ Review?\end{tabular} &
  ML? &
  DL? &
  NLP? &
  EHR? &
  \begin{tabular}[c]{@{}c@{}}Evaluation\\ Metrics?\end{tabular} &
  \begin{tabular}[c]{@{}c@{}}Word\\ Embeddings?\end{tabular} &
  \begin{tabular}[c]{@{}c@{}}Feature\\ Extraction?\end{tabular} &
  \begin{tabular}[c]{@{}c@{}}Clinical \\ Tools?\end{tabular} &
  \begin{tabular}[c]{@{}c@{}}Transformer\\ Model?\end{tabular} \\ \hline
2022 & This paper           & $\sqrt{ }$ & $\sqrt{ }$ & $\sqrt{ }$ & $\sqrt{ }$ & $\sqrt{ }$ & $\sqrt{ }$ & $\sqrt{ }$ & $\sqrt{ }$ & $\sqrt{ }$ & $\sqrt{ }$ \\ \hline
2022 & Tyagi et al. \cite{tyagi2022neurahealthnlp}          & $\chi$  & $\sqrt{ }$ & $\sqrt{ }$ & $\sqrt{ }$ & $\sqrt{ }$ & $\sqrt{ }$ & $\chi$  & $\chi$  & $\chi$  & $\sqrt{ }$ \\ \hline
2021 & Chowdhury et al. \cite{chowdhury2021use}      & $\sqrt{ }$ & $\sqrt{ }$ & $\sqrt{ }$ & $\sqrt{ }$ & $\sqrt{ }$ & $\chi$  & $\chi$  & $\chi$  & $\chi$  & $\chi$  \\ \hline
2020 & Juhn et al. \cite{juhn2020artificial}           & $\chi$  & $\chi$  & $\chi$  & $\sqrt{ }$ & $\sqrt{ }$ & $\chi$  & $\chi$  & $\chi$  & $\chi$  & $\chi$  \\ \hline
2020 & Ahmed et al. \cite{ahmed2020identification}          & $\chi$  & $\chi$  & $\sqrt{ }$ & $\sqrt{ }$ & $\sqrt{ }$ & $\sqrt{ }$ & $\chi$  & $\chi$  & $\chi$  & $\chi$  \\ \hline
2020 & Wu et al. \cite{Wu2020Deep}             & $\sqrt{ }$ & $\chi$  & $\sqrt{ }$ & $\sqrt{ }$ & $\sqrt{ }$ & $\chi$  & $\sqrt{ }$ & $\chi$  & $\chi$  & $\sqrt{ }$ \\ \hline
2019 & Alzoubi et al. \cite{alzoubi2019review}        & $\chi$  & $\sqrt{ }$ & $\chi$  & $\sqrt{ }$ & $\sqrt{ }$ & $\sqrt{ }$ & $\chi$  & $\sqrt{ }$ & $\chi$  & $\chi$  \\ \hline
2019 & Juhn et al. \cite{juhn2019natural}           & $\chi$  & $\chi$  & $\chi$  & $\sqrt{ }$ & $\sqrt{ }$ & $\chi$  & $\chi$  & $\chi$  & $\chi$  & $\chi$  \\ \hline
2019 & Koleck et al. \cite{koleck2019natural}    & $\chi$ & $\chi$  & $\chi$  & $\sqrt{ }$ & $\sqrt{ }$ & $\chi$  & $\chi$  & $\chi$  & $\chi$  & $\chi$  \\ \hline
2018 & Wang et al. \cite{wang2018clinical} & $\chi$  & $\chi$  & $\chi$  & $\sqrt{ }$ & $\sqrt{ }$ & $\chi$  & $\chi$  & $\chi$  & $\chi$  & $\chi$  \\ \hline
2017 & Luo et al. \cite{luo2017natural}   & $\sqrt{ }$ & $\chi$  & $\chi$  & $\sqrt{ }$ & $\sqrt{ }$ & $\chi$  & $\chi$  & $\chi$  & $\chi$  & $\chi$  \\ \hline
\end{tabular}%
}
\end{table}

This paper makes a significant contribution in that by covering a comprehensive systematic review that fills a gap in the existing research. We focus on (1) the commonly utilised ML and DL-based models, including their importance in healthcare NLP; (2) the popular ML and DL models with their feature extraction or word embedding and evaluation matrix; (3) various applications of NLP, including transformer model, applied in the EHRs; (4) commonly used data types, clinical free text preprocessing pipeline and study settings; and (5) existing automated ML-enabled tools used by health professionals and healthcare industries. We further highlight the core challenges of medical NLP, the trend of current research, and the shortcomings of the existing literature. We end this work by addressing our review's findings and highlighting study shortcomings and future goals. 

The remaining sections of the review article are structured as follows. We explain the literature search and selection strategy in Section \ref{search_strategy}. The techniques used for analysing EHR are illustrated in Section \ref{adopted_models}. Discussions and research viewpoints are delineated in Sections \ref{discussions} and \ref{research_view}. Finally, the paper is concluded in Section \ref{conclusion}.

\section{Methodology}

\subsection{Literature search and selection strategy}
\label{search_strategy}
We searched eleven electronic databases from 2016 to 2022. Significant contributions have been made to NLP research in the last six years \cite{jabali2022electronic}. This search was done through well-established outlets hosting a wide range of high-quality peer-reviewed articles. We have searched Google Scholar, PubMed, Elsevier, IEEE, Springer, Oxford University Press, Nature Publishing Group, Wiley Online Library, BioMed Central, and American Medical Informatics Association for literature from 2016-2022. These databases are rich and have high-quality peer-reviewed articles for NLP research. We have developed several key terms to identify the studies, like “NLP in Clinical Narratives”, “Medical NLP”, “ML in EHRs”, “DL in Medical Text”, “Automated ML in EHRs”.

\subsection{Selection criteria}

\textbf{A) Inclusion Criteria:} We have included literature that described: ML/DL-based free text classification, word-embedding approach in the context of medical text data, automatic clinical narratives summarisation, healthcare dialogue system, medical concept embedding, Delirium risk identification, ICD-9 multi-label classification, clinical entity recognition, machine learning and deep learning architecture for EHRs.  Only peer-reviewed journal articles or full conference papers were included.  To be included, a study must have used an ML or DL-based model or framework designed solely for analysing EHRs. Studies must also have been focused on analysing and identifying clinical narratives through ML or DL methods.

\textbf{B) Exclusion Criteria:} Research works that were published as a preprint, with preliminary work or without peer review, were excluded. Editorials and review papers were also on the exclusion list. After the initial screening, the articles' retrieved for full-text analysis also was examined for quality.

\subsection{Search Output:} 
\begin{figure}[!ht]
    \centering
    \captionsetup{justification=centering}
    \includegraphics[width=0.7\textwidth]{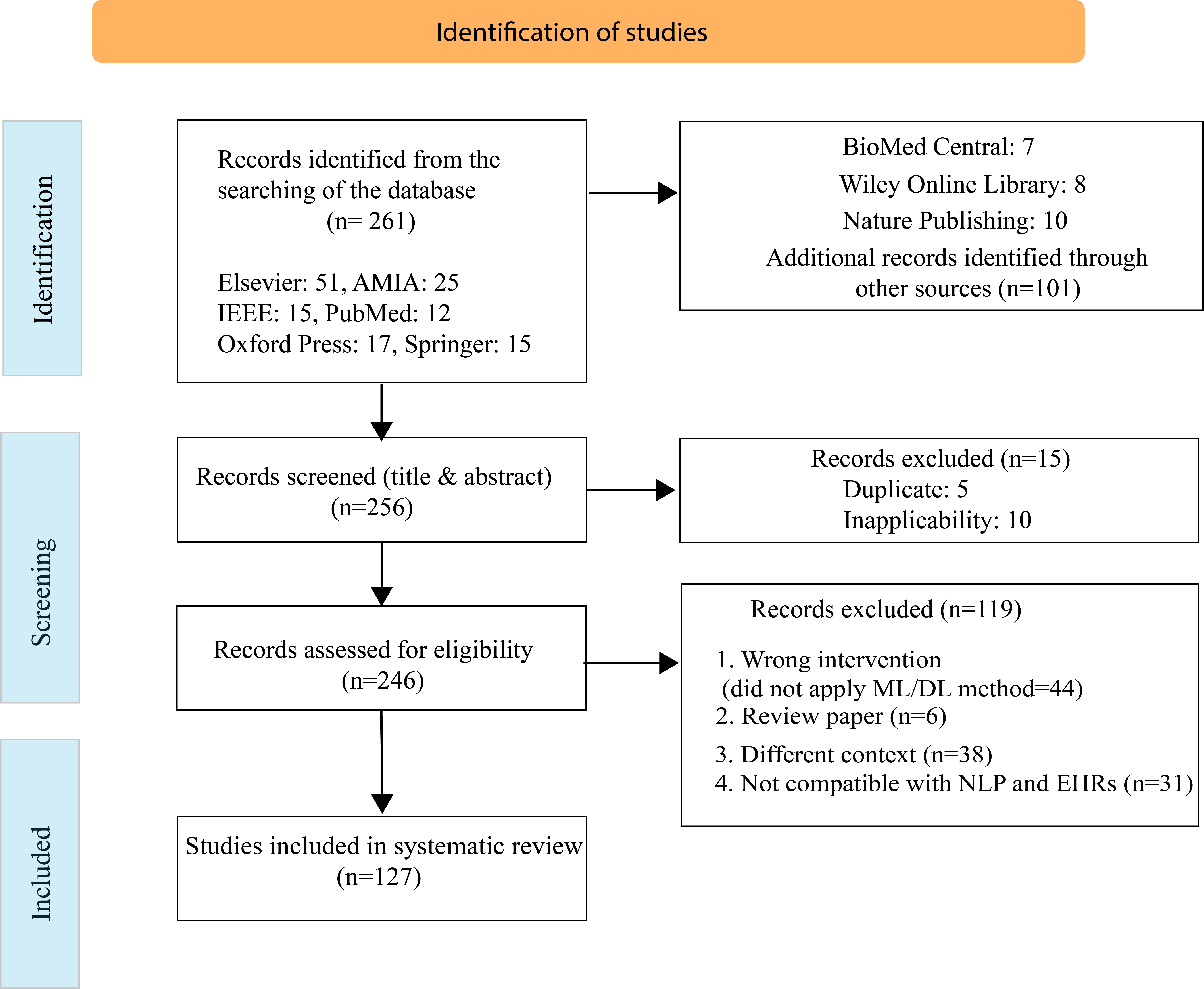}
    \caption{Literature exclusion and inclusion results were followed by the PRISMA method. }
    
    \label{fig:prisma_method}
\end{figure}

Figure~\ref{fig:prisma_method} illustrates that in the initial search, $261$ titles were identified for the title and abstract screening, comprising $15$ from Springer, $12$ from PubMed, $15$ from IEEE, $51$ from Elsevier, $17$ from Oxford University Press, $10$ from Nature Publishing Group, $25$ from American Medical Informatics Association (AMIA), $7$ from BioMed Central, $8$ from Wiley Online Library. Of these, $119$ papers were excluded based on our exclusion criteria. An additional $101$ papers were identified from the reference lists of retrieved articles. Four of these titles were duplicates and thus excluded, $1$ article was not available for review, and ten did not match our criteria. The remaining $127$ papers were retrieved for full-text review.

\begin{figure}[!h]
    \centering
    \captionsetup{justification=centering}
    \includegraphics[width=0.6\textwidth]{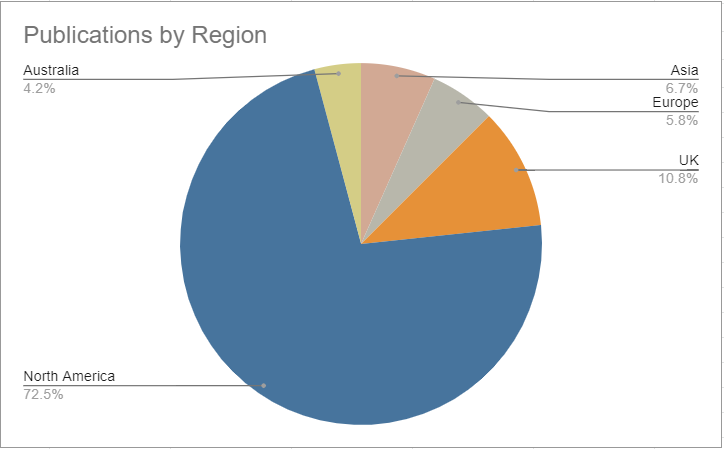}
    \caption{Medical NLP research around the globe.}
    
    \label{fig:std_charc}
\end{figure}

Studies included in the review described different traditional and hybrid methods for analysing clinical free text. Some proposed new approaches and others tried to improve existing techniques. The majority of studies originated from North America, followed by the United Kingdom, Asia, Europe and Australia, as illustrated in Figure~\ref{fig:std_charc}. No articles from the African continent or South America met our criteria. The sample size used in the retrieved articles varied between $150$ and $823,627$. EHR was the most used data type and was used as the sole data source by almost all studies. Most datasets were unstructured, and two studies only used structured free-text data. The study designs were mostly experimental ($n=24$), cohort ($n=9$), case study ($n=1$). Among the existing articles we reviewed, some studies ($n=11$) used diagnostic tools without specifying the name of the diagnostic tool, while others ($n=9$) explicitly stated that they used the International Classification of Diseases (ICD) tool, which is used to classify disease and mortality codes.

 Table~\ref{tab:table_data_type}.

\section{Techniques used in the literature for Analysing EHR}

\label{adopted_models}
Techniques used in EHR can be categorised into the following groups: patient risk analysis/prediction, advanced architectures to analyse EHRs, medical text summarising and other NLP applications.

\subsection{Patient Risk Analysis/Prediction}
This section presents an overview of the essential concepts of machine learning and natural language processing applied to analyse and predict the risk condition of patients. We will discuss articles concerning patient risk analysis and prediction, including 1) perinatal self-harm detection, 2) suicide attempt detection, 3) automated HIV risk assessment, 4) delirium detection, and finally, 5) diagnosing lupus nephritis.

\textbf{A) Perinatal Self-harm}

Mental illness, drug abuse, singleness, and obstetric and neonatal complications are major risk factors for self-harm throughout pregnancies and the first year following delivery  \cite{ayre2019prevalence}, which is likely to be the case of perinatal self-harm. It is worthwhile to note that NLP has been used to identify suicidality in EHRs \cite{bittar2019text}, including those of kids suffering from autism spectrum disorders  \cite{downs2017detection} and primary care physicians  \cite{anderson2015monitoring}. 

Several studies have focused on women with Serious Mental Illness (SMI) during the perinatal stage. 
Ayre et al. \cite{ayre2021developing} proposed a Natural Language Processing (NLP) tool that can effectively identify those who have perinatal self-harmed. Clinical Record Interactive Search (CRIS) was used in this study, which enabled researchers to access women's de-identified medical health records. The tool investigated by Ayre takes a text as input ("She took an overdose", "Previous episodes of self-harm", "Current episode of self-harm") and sequentially runs it through five processing layers before generating an Extensible Markup Language (XML) file in which XML tags annotate each detected instance of self-harm and its associated attributes. The processing layer includes linguistic preprocessing, lexical rules, token sequence rules, negation detection and contextual search. The developed tool was validated through precision, recall and f1-score, and the tool's performance in detecting perinatal self-harm was found satisfactory. However, there remain several shortcomings in this study. The sample size of the dataset was relatively small, which may create the possibility of overfitting problems. Overfitting occurs when a model acquires so much information and noise from the training data that it impairs its performance on new data. In addition, this tool is currently running as a beta version, and full development is not yet complete making it difficult to measure efficiency.

\vspace{5mm}

\textbf{B) Identify Suicide Attempts}

Identifying first-time suicide attempts has always been challenging since prediction models generally demand huge data sets \cite{belsher2019prediction}. In addition, risk assessment mainly depends on patient-reported data \cite{simon2018predicting} and, and patients may be prone to hide suicidal notions \cite{walsh2018predicting}. These have hampered the accurate identification of suicide risk over the years. In light of the underlined context, a group of researchers \cite{tsui2021natural} conducted a study using electronic health records to detect patients who are at-risk for their first attempt at suicide using machine learning and natural language processing.

An open-source NLP tool, "cTAKES", was used to extract clinical outcomes from medical notes, requiring no preprocessing. This tool has been widely used and thoroughly tested to process numerous descriptive notes, including discharge summaries, radiology notes, history and physical progress. When extracting clinical concepts from large-scale medical records, Concept Unique Identifiers (CUIs) from the Unified Medical Language System (UMLS) was used to annotate each concept. Furthermore, Tsui et al. \cite{tsui2021natural} employed traditional machine learning models to predict the suicide risk by exploiting the retrieved features of EHRs: Random Forest (RF), Least Absolute Shrinkage and Selection (LASSO) regression, Naïve Bayes (NB),  as well as the Ensemble of Extreme Gradient Boosting (EXGB). This study adopts 3 frameworks based on feature engineering concerning feature optimisation: wrapper, filter and embedded. The main idea of this optimisation is to reduce the number of input variables in order to lower the computational cost of modelling, thus improving the predictive model's performance. Note that the proposed algorithms were not compared with more established models for predicting suicide, making it hard to gauge the improvement it offers.

\begin{figure}[H]
    \centering
    \captionsetup{justification=centering}
    \includegraphics[width=0.9\textwidth]{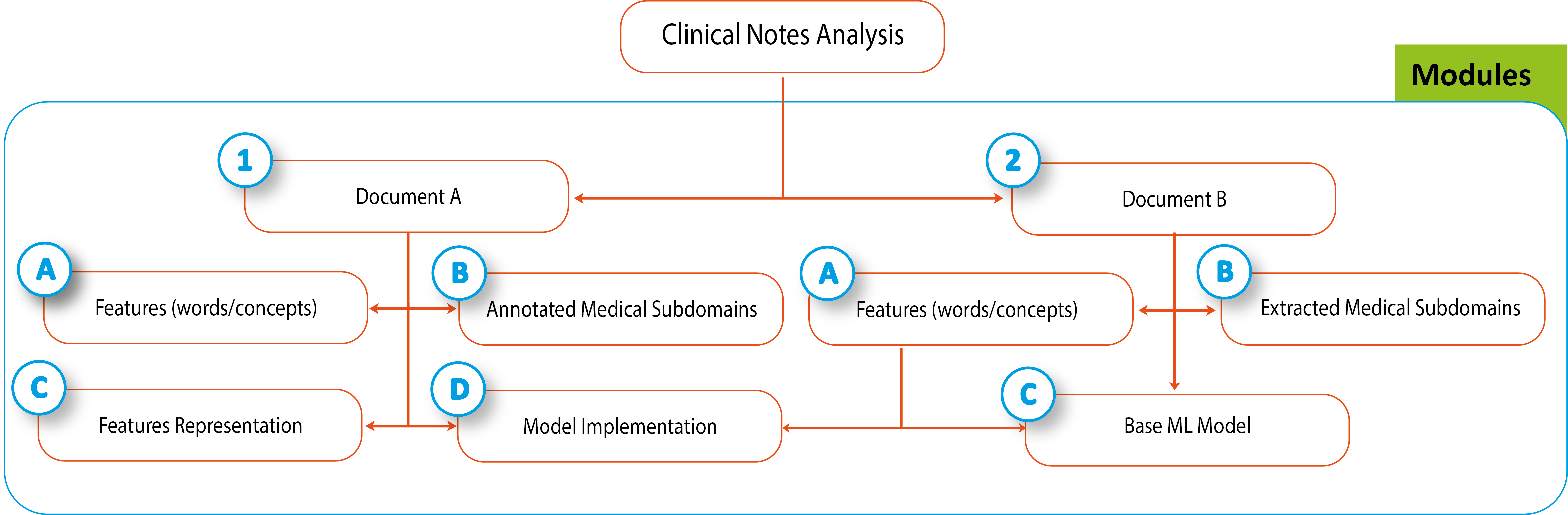}
    \caption{Clinical notes analysis's architecture diagram.}
    
    \label{fig:clinical_note}
\end{figure}

Using a different approach, Carson et al. \cite{carson2019identification} developed and evaluated an ML method using NLP in EHR to detect suicide in adolescents. In order to categorise teenagers according to their history of suicide attempts, the authors illustrate the implementation of an ML system that creates a classification model from codes created by NLP analysis of EHRs notes. In this work, Invenio software was employed to encode the unstructured textual information of EHRs. Invenio is built upon an open-source Apache cTAKES platform \cite{savova2010mayo} and analyses unstructured descriptions of medical notes. Compared to the cTAKES system, Invenio was found to be more successful when converting free clinical text; More specifically, this system uses several features such as a sentence boundary detector, tokeniser, normaliser and part-of-speech tagger, shallow parser, and named entity recognition annotation. In addition, Invenio's performance in capturing negative phrases in electronic health records proved to be satisfactory. Furthermore, a random forest (RF) classifier was also applied to classify individual patients according to the history of prior suicide attempts. Again, the authors used the five-fold cross-validation technique to optimise the features of the proposed model. Finally, Area Under the Curve (AUC) was used to assess the performance of the model.

\vspace{5mm}

\textbf{C) Automated HIV Risk Assessment}

Many studies have proposed solutions for automated Human Immunodeficiency Virus (HIV) risk assessment. These studies primarily rely on structured medical free text and have many limitations in capturing important information on HIV risk factors. Usually, narrative or semi-narrative formats are used to gather precise descriptions of social and behavioural factors, such as sexual orientation and sexual activity.

Utilizing machine learning and natural language processing, Feller et al. \cite{feller2018using} suggested a strategy to detect persons at high risk for HIV using EHRs. This automated diagnosis was carried out in four steps: keyword identification, topic modelling, variable selection and statistic modelling. The first stage aims to identify words with potentially rich information value by representing each word in the clinical note based on its Term Frequency-Inverse Document Frequency (TF-IDF) weight. The second stage focuses on topic modelling, from which large amounts of text can be analysed, and its content can be defined by focusing on hidden features with a certain weight. This was done using the Latent Dirichlet Allocation (LDA) algorithm, which takes a corpus of notes as input and learns K clusters, representing the distribution of words in each corpus. The third step is essential when working with a large number of clinical notes. It can be said that redundant variables often reduce the performance of predictive models, so it is recommended to eliminate irrelevant features and keep the corpus simple before feeding the data to the ML system. Hence, Feller et al. identified a selection of the most valuable variables using mutual information criteria. The relationship between two random variables is quantified by mutual information, which may account for both linear and non-linear correlations.

The final step illustrates the development and assessment of the statistical model. Feller et al. utilised random forest classifiers to predict the risk factors of HIV as they are simple to tune and provide a measure of variable importance; following the bagging method reduces the chance of being aﬀected by outliers and thus allows interpretation. Consequently, the suggested model was evaluated through the model validation indicators such as precision, recall and f1-score.

\textbf{D) Delirium Identification}

Delirium is a sudden onset of confusion that can last for hours or even days. Delirium is often not classified for billing and is underdiagnosed in clinical practice. Although manual chart review can be employed to detect the presence of delirium, it is time-consuming and inappropriate for large-scale investigations. Since NLP has the ability to process and determine raw text data, Fu et al. \cite{fu2022ascertainment} were motivated to implement and evaluate NLP algorithms to detect delirium events from EHRs.

The authors developed two NLP techniques such as NLP-CAM and NLP-mCAM, using the Confusion Assessment Method (CAM). CAM is a standardised, evidence-based method that allows physicians without psychiatric training to effectively identify delirium in clinical and research contexts. The NLP models examine patient charts for clear indications of delirium patients and its associated medical information that fits the CAM standards to determine an individual's delirium condition. The CAM has four features that help assess delirium: sudden onset and variable course, lack of attention, disordered thoughts, and altered level of awareness. Each delirium and its associated concept were normalised to the correct format based on these features. In addition, any erroneous examples found during training were inspected through a manual process and repeatedly corrected until all errors were rectified, paving the way for improvement of the model. Thus, the proposed NLP techniques have worked admirably to identify patients experiencing delirium using health records in a timely and cost-effective manner.

\vspace{5mm}

\textbf{E) Identify Lupus Nephritis}

Lupus nephritis is a kind of kidney condition that is triggered by Systemic Lupus Erythematosus (SLE or lupus). Much of the information needed to detect Systematic Lupus Erythematosus (SLE or lupus), such as histology notes for kidney biopsies, are only available in text-based notes, making it difficult analyse rule-based detection algorithms and text string searches. Researchers have come up with innovative solutions to diagnose lupus nephritis, but their solutions are not very effective. A study conducted by Deng et al. \cite{deng2021natural} designed an NLP system to analyse the clinical notes to detect the early onset of nephritis. In this study, the authors utilised two inpatient and outpatient datasets and implemented 4 algorithms: a rule-based algorithm that utilises only structured data (baseline algorithm) and other 3 algorithms utilising various NLP-based models. Each of the 3 NLP models is built on l2-regularized logistic regression utilizing a separate feature set, comprising positive mention of Concept Unique Identifiers (CUIs), number of occurrences of CUIs, and a blend of all three components, respectively. 

Furthermore, Deng et al. preprocessed the medical records by removing identical entries and lemmatisation phrases. The MetaMap was used to tag medical terms within these phrases. In addition, the SHAP decision plot was also applied for assessing feature relevance to present which features were more important during diagnosis. Again, the proposed algorithms were compared with the three independent NLP models and a baseline algorithm to ensure that this method could efficiently identify individuals with lupus nephritis. Nonetheless, the suggested approach makes it easier to accurately diagnose this disease, which helps researchers better understand the SLE characteristics of individuals. Yet, missing laboratory tests from EHRs with this small sample size (50) affected prediction accuracy.

\subsection{Advanced Architectures to Analyse EHRs}
Deep learning and transfer learning architectures have revolutionised clinical NLP in recent times. The recent results of several pre-trained models against known benchmarks solidify transfer learning's position as an indispensable technique in modern NLP. More specifically, the state-of-the-art DL and transformer-based model have been used in various NLP tasks over the past years. Examples include, automated ICD-9 coding, multi-classification problems, medical text summarisation, language translation, clinical data de-identification, etc. The following subsection summarises deep learning and transfer learning architectures frequently applied to analyse EHRs. 

\vspace{5mm}

\textbf{A) Convolutional Neural Networks} 

DL methods are beginning to lead a wide range of clinical NLP applications due to their low complexity, fast processing, and state-of-the-art results in the automated the International Classification of Diseases (ICD-9) classification. Li et al. \cite{li2018automated} designed a deep learning system (DeepLabeler) to classify ICD-9 automatically. This framework consists of Convolutional Neural Network (CNN) with Document to Vector (D2V) technique to retrieve and encode local and global features. The proposed model performs its task by following two steps: 1) feature extraction and 2) multi-label classification. In the feature extraction phase, Li et al. effectively extracted global and local features from the Medical Information Mart for Intensive Care (MIMIC) dataset with the recent success of D2V techniques. Li et al. demonstrated that the adopted D2V approach keeps all words in one document for training; Thus, it does not eliminate any useful information. Compared to the CNN model, it is quite impossible to retain the full words in the document during training because it considers ignoring semantic information while extracting features. Secondly, the multi-label classification steps utilise a Fully Connected Neural Network (FCNN), sigmoid activation function and backpropagation technique to anticipate the likelihood of each ICD-9 code. On the other hand, due to the small number of documents used in this study \cite{li2018automated}, it is noticeable that the F-measure was not exceptionally high. The imbalanced distribution of ICD-9 codes in the MIMIC dataset is primarily the reason for the low F-measure in ICD-9 automatic coding.

In addition to the wide variety of clinical NLP applications, researchers do not limit themselves to contemporary NLP solutions, as the research community continuously strives to solve several significant problems in the current NLP space. It can be said that among the existing articles, clinical name entity recognition (NER) is not widely available; However, recently, a customised NER model has been adopted to extract several medical entities from a large number of medical records. Kormilitzin et al. \cite{kormilitzin2021med7} proposed a NER model to identify clinical entities in seven categories: Drug Name, Route of Administration, Frequency, Dosage, Strength, Form and Duration. The recommended model was developed based on the spaCY python-based open source library. While several excellent libraries are available in full versions, including NLTK \cite{bird2009natural}, Stanford CoreNLP \cite{manning2014stanford}, Hugging Face \cite{wolf2020transformers}, and NLP4J \cite{choi2016dynamic}, the Spacy library is optimised for CPU speed. The core architecture of the proposed NER model is based on a CNN network. The token representations are hashed Bloom embeddings of specific word prefixes, suffixes and lemmatisations complemented by a transition-based chunking model. In the case of model performance, in seven areas, it received a micro-mean F1 score of 0.957. In addition, the transferability of the created model was evaluated utilising data from the United Kingdom Secondary Care Mental Health Record (CRIS) from United States critical care facilities.

\textbf{B) Long Short-Term Memory} 

A hybrid model of Gated Attention incorporated Bi-Directional Long Short-Term Memory (ABLSTM), and attention-based bi-directional LSTM was proposed by Li et al. to classify clinical text \cite{li2021hybrid}. In this study, Li et al. applied a three-stage hybrid system that incorporates the threshold-gated neural network model with the attention-guided rule-based approach to solve a multi-class clinical text classification problem. To begin with, Recurrent Neural Network (RNN) was applied in this study to be effective for modelling time-sensitive sequences. On the other hand, the fundamental concept behind LSTM was to implement "gates" to regulate the data flow to RNN units~\cite{haque2018image}. Furthermore, the attentive recurrent architecture was introduced because Li et al. observed that when dealing with medical multi-class classification problems, the notable drawbacks of "black box" methods cannot be ignored. In order to overcome this problem, the authors included a bi-directional LSTM framework, including an attention layer to enable the network to weigh the words in a phrase based on their perceived relevance. Besides, the weighted average and occurrence filter method was prioritised for calculating word weight. Finally, a three-stage hybrid method was developed that applied three subsequent modules to obtain the final output: GATED ABLSTM classifier, regular expression-based classifier, and ABLSTM classifier. However, Li et al. found that the traditional LSTM network limited the ability to receive significant scores for a particular word in the input document.

\begin{figure}[H]
    \centering
    \captionsetup{justification=centering}
    \includegraphics[width=0.8\textwidth]{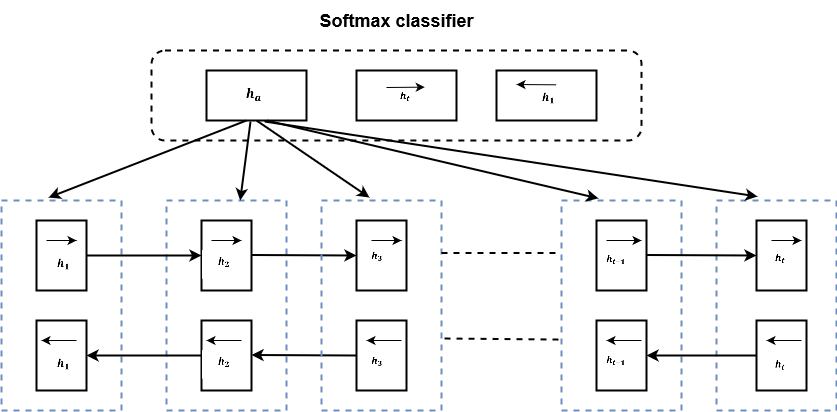}
    \caption{Architecture of the attention-based bi-directional LSTM.}
    \label{fig: attention_architecture}
    
\end{figure}

\vspace{9mm}

\textbf{C) Recurrent Neural Networks} 

Another pioneering effort has shown remarkable achievement in the context of multi-class classification problems. Recent breakthroughs in advanced deep learning models show great utility in the NLP space. A hybrid deep learning model was proposed in \cite{nigam2016applying} to classify ICD-9 codes toward multi-class classification. This study applied RNN, LSTM, baseline logistic regression, feed-forward neural network and Gated Recurrent Unit (GRU) as part of the hybrid approach. Nigam et al. collected the Medical Information Mart for Intensive Care (MIMIC III) dataset, consisting of de-identified medical records and various clinical abbreviations, common misspellings, clinical phrases, and so on. This study preprocessed noise and less important information from the dataset in the first phase to create a clean corpus. Secondly, creating vocabulary was a primary concern when selecting features. Note that abbreviations, misspellings and idiosyncrasies were ignored during feature selection as they were less significant. Nigam et al. applied the Bag of Words (BOW) approach to extract features from the text document. In the case of model selection, first, the baseline Logistic Regression (LR) model was applied to train a separate model for each class, and each model accurately forecasted the predicted value (0 or 1). Afterwards, the RNN model was created, and this model performed differently than the baseline LR model. For example, instead of summing the entire BOW note vector, each vector was separated and input a normalised vector at each time step. To avoid losing semantic information from previous notes, the labels of the notes were replaced with LSTM units. Although the proposed model successfully classified ICD-9 codes, there is still a limitation. The model appears to be overfitting the training data and probably requires a higher dropout value.

\textbf{D) Transformer-based Architectures}
Recent advances in transfer learning have gained popularity in clinical NLP. More specifically, the Bidirectional Encoder Representation from Transformer (BERT) combines bidirectional transformers and transfer learning to create state-of-the-art models for various NLP tasks \cite{kici2021bert}. In recent years, Mulyar et al. \cite{mulyar2021mt} proposed a Multitask-Clinical BERT (MT-Clinical BERT), which is a unique model that combines individual activities. . For example, it conducts multi-task learning on 8 different information retrieval tasks, including entity retrieval and identification of Personal Health Indicator (PHI), in addition to clinical text embedding learning. These embeddings are fed as input to these prediction functions. This multi-task strategy is competitive against task-specific information extraction algorithms due to its capacity to exchange data across several inconsistently annotated datasets.

Another transformer model called BEHRT was suggested by Rao et al. \cite{li2020behrt}, which learns about the previous illnesses of patients as well as the interconnections between them. This algorithm is specifically designed to forecast a patient's future diagnosis (if any), given their past symptoms. The proposed BEHRT creates a definitive embedding through information on disease progression and care delivery as well as maintaining event timing. Compared to previous techniques such as RETAIN \cite{liu2018learning}, an assessment of the model revealed that BEHRT had greater predictive power, as shown by an increase of 8.0–13.2\% in average accuracy scores for tasks such as sickness trajectories and illness prediction.

In another case, it is often seen that actual clinical data is underutilised in most studies because researchers often do not have access to actual data due to data scarcity and confidentiality. Note that most studies focused primarily on using the MIMIC corpus; nevertheless, MS-BERT was created by Costa et al. \cite{costa2020multiple} as the first publicly accessible transformer model that was trained on actual clinical data. The MS-BERT model is publicly accessible and has been trained on more than 70,000 consultation notes for Multiple Sclerosis (MS) patients. It is to be mentioned that the notes were de-identified before training. Furthermore, the model was evaluated using a classification task in order to forecast the Expanded Disability Status Scale (EDSS). In the macro-F1 score, the model outperforms competing models using word2vec, CNN and rule-based techniques.

In addition, a group of researchers \cite{smit2020chexbert} developed CheXbert, which employs BERT to classify free text radiological records. Existing machine learning models in this study use feature engineering or human annotation. Although of excellent quality, annotations are limited, and production is expensive. The CheXbert overcomes this issue by learning to classify radiography reports via annotations and current rule-based techniques. It first learns to anticipate the output of a rule-based labeller, then fine-tunes an extensive set of expert comments. It achieved a new state-of-the-art result by improving F1 scores for a report labelling task on the MIMIC-CXR dataset \cite{johnson2019mimic}, which contains large-scale labelled chest radiographs. 

\subsection{Medical Text Summarising}

Creating a summary system from medical narratives has become challenging as few effective tools have been developed in the healthcare sector. When large amounts of data are collected, such as in the Intensive Care Unit (ICU), displaying efficient data becomes a key concern for strategic planning. While the most common strategy would be visually displaying information, text summarisation has already been found to aid strategic planning. This section will review the abstract and extractive text summarisation approaches researchers have developed in the retrieved articles.

\textbf{A) Extractive Summarisation Models} 

Portet et al. \cite{portet2009automatic} proposed an extractive summarisation model called BT-45 utilising EHR to generate textual summaries of approximately forty-five minutes of constant clinical information and random events. This study developed text summaries in four steps, each accessing area-specific knowledge containing conceptual content in neonatal intensive care. Signal analysis (1), which extracts the basic characteristics of the physiological time-series data, is the first phase of processing (artefact, pattern and trend). In order to comprehend more creative clinical findings and linkages derived from data pertaining to signal features and random occurrences, data interpretation (2) employs a variety of time and rational reasonings. The third phase of documentation planning (3) organizes the most relevant occurrences from the preceding phases into a tree of related occurrences. Eventually, this tree is transformed into a coherent text through microplanning and realisation. However, human evaluators find model summaries ineﬃcient because of the diﬃculty in integrating these disparate data into one model. As a result, most of the following initiatives focus primarily on textual information. Despite these limitations, the authors demonstrate that it is feasible to construct concise paragraphs from huge, complicated information that may be used as useful planning instruments.

Moradi et al. \cite{moradi2019small} introduced a graph-based algorithm that analysed words and phrases using biomedical text. In the first phase, the primary concern was drawing out technical reports' content. This task was accomplished per the document's formatting and logical structure. In the case of scientific articles, the main body was extracted by eliminating those parts of the text that seemed unnecessary to include in the abstract. The title, author information, abstract, keywords, section and subsection headings, bibliography, and other elements were included in these parts. Subsequently, Moradi et al. use the Helmholtz principle from Gestalt theory to determine the concepts that convey primary information from the text and then construct a graph from that information. Finally, the degree of each node in that generated graph was calculated by a summation, which then ordered the nodes in descending order. In addition, the ROUGE score was used to evaluate the proposed model, which calculates different ratings indicating the content similarity between a reference summary and the summary made by a machine learning model. Also, a comparative analysis of this method with other summaries was carried out during the evaluation period. This model had the best ROGUE value among other comparison techniques. However, the authors intend to extend this research by using increasingly advanced methods at different stages of the summarisation process.

Another extractive summarisation model developed by McInerney et al. \cite{mcinerney2020query} selects sentences most likely related to a potential diagnosis. First, Mclnerney et al. compiled a list of individual reporting forms and diagnostic codes from each individual, chronologically arranged by date and time in various collected EHR datasets. Then, their strategy was to train a deep learning model incorporating a transformer-based approach to select short phrases from EHRs. For this, the authors developed and evaluated systems for remote supervision, which only require grouping diagnostic codes. Such systems employ medical BERTs to encode questions and comments simultaneously and then identify groups of ICD codes that correlate with specific illness diagnoses used to train the model. Although the proposed remotely supervised model significantly outperforms unsupervised baseline models, Mclnerney et al. intend to expand this research to determine whether adding a little direct supervision can improve the model's performance further.

\textbf{B) Abstractive Summarisation Model}

Radiological reports have been increasingly summarised using seq2seq and related models in abstractive summarisation research. The first study using seq2seq to develop radiological impressions was conducted by Zhang et al. \cite{zhang2018learning}. Zhang et al. suggest using the neural seq2seq method for making radiological assessments. The authors also propose a particular deep learning model for this activity that learns to encode prior research knowledge and uses it to guide the decoder. Additionally, a pointer-generator model was used in the decoding part. Consequently, this model outperforms state-of-the-art baseline models on larger datasets of radiological records collected from real hospital trials using the ROUGE system. Although the background part of the report was simplified by encoding the model as an abstract summary, radiology specialists' treatments are often excluded from the results part and require an extensive understanding of research and field expertise; the model often misses follow-up treatments.

\subsection{Other NLP Applications}

The additional applications of NLP can be clustered under the following headings: blockchain-based EHRs, identifying goals-of-care conversations, clinical chart review and medical language translation. 

\textbf{A) Blockchain-based EHRs} 

The significance of having a reliable record-tracking and communications mechanism has been highlighted more recently worldwide during the COVID-19, which indicates the current inadequacy in this field. Bharimalla et al. \cite{bharimalla2021blockchain} focused on a blockchain and NLP-based approach to make a communication and record tracking system [33]. The proposed prototype system is based on Hyperledger Fabric, a distributed ledger technology that is open source and designed for enterprise use. It is a popular choice for private blockchains. Bharimalla et al. categorised their methodology into system architecture, data pulling and sharing, patient data management and converting paper prescriptions to text. To be more specific, Bharimalla et al. focused on converting paper prescriptions into text using NLP methods to integrate old paper-based clinical records into the new system using a mobile application-based interface. 

\begin{figure}[H]
    \centering
    \captionsetup{justification=centering}
    \includegraphics[width=0.7\textwidth]{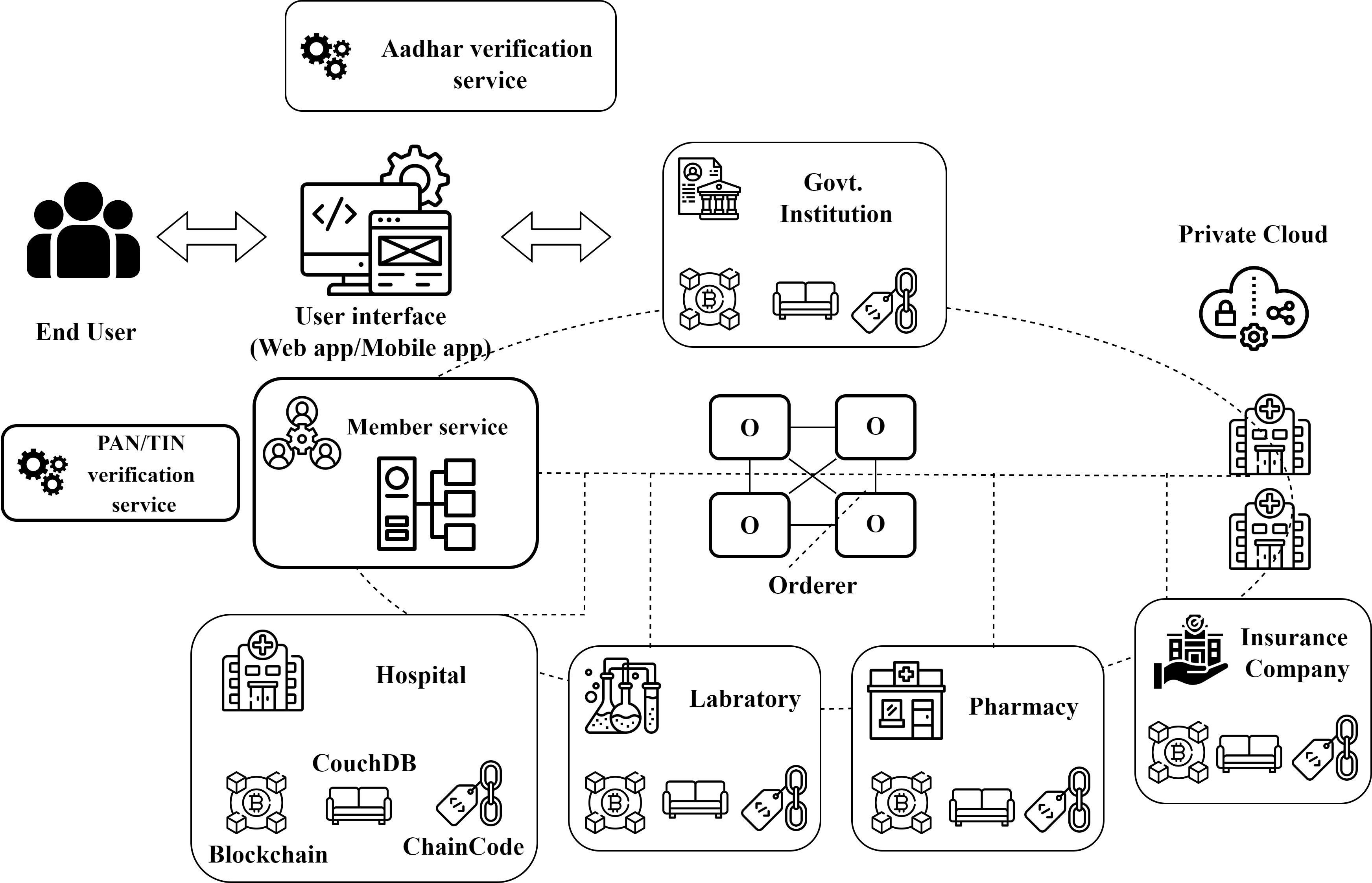}
    \caption{Framework for Blockchain-based Healthcare System \cite{bharimalla2021blockchain} . }
    
    \label{fig:blockchain}
    
\end{figure}

Turning to the data extraction phase, CNN, LSTM and Residual Networks (ResNet) were applied in terms of extracting handwritten data. At the same time, the Google tesseract model was considered to extract printed prescriptions data. Moreover, Bharimalla et al. carried out some preprocessing in the extraction process. Firstly, the image was converted to grayscale to create a more functional model. The next step was to apply the Otsu thresholding process, which turns the pixels into ones and zeros after the grayscale operation. Note that some pixels are usually lost during thresholding; therefore, to mitigate these problems, Erosion and Dilation techniques were used to restore certain pixels, where Erosion enlarges some pixels and Dilation reduces some pixels. Bharimalla et al. explain that these images will be forwarded to Google-Tesseract after completing preprocessing. Finally, all text from the image will be extracted by tesseract and sent over the network. Figure \ref{fig:blockchain} illustrates the proposed framework for a blockchain-based healthcare system. It shows the main participants, elements, and transaction procedures.

\textbf{B) Identify Goals of Care Conversations}
Goals-of-care conversations aid patients with severe illnesses in articulating what they value most and wish to occur with their medical treatment. Medical professionals can use this information to create a care plan based on the patient's values and preferences. In light of the context, Lee et al. \cite{lee2021identifying} developed an automated method for identifying goals of goal-of-care discussion using NLP approaches. In brief, a sample of $3183$ EHR notes was collected from $1426$ patients with severe illnesses, and each note was manually evaluated for documentation of goals-care discussions. The EHR notes were randomly divided into 100 training and test set pairs. The NLP technique was used to tokenise each note in unigram (i.e., one-word length tokens), removing common stop words and negation terms. In this study, the logistic regression classifier was applied for each training set and measured the classifier's performance using the Area under the receiver operating curve (AUC). The authors divided the data samples into inpatient or outpatient datasets and used the same methodology for training and testing the model in both subgroups to investigate the suggested model's effectiveness. However, Lee et al. stated that additional research is required to validate the proposed approach, particularly in identifying outpatients' goals-of-care discussions.

\textbf{C) Clinical Chart Review} 

Periprosthetic joint infection (PJI) data elements exist in both unstructured and structured EHR records and must be collected manually. This study \cite{fu2021automated} aims to create an NLP technique to simulate manually annotated chart evaluation for data items of PJI. The suggested strategy was based on expert rules that focused on textual cues (i.e., PJI-related terms) identified in orthopaedic surgeons' or communicable diseases experts' clinical narration. The text preprocessing, concept extraction, and classification processes are the three primary parts of the NLP method. Sentence segmentation, assertion detection, and temporal extraction were the critical elements of the textual data processing workflow. Additionally, concept extracting is a knowledge-driven annotating and indexing technique that recognizes phrases in the unstructured text that correspond to topics of interest. Furthermore, in developing the NLP algorithm, a training sample of 1208 TJA surgeries (170 PJI cases) and a test sample of 1179 TJA surgeries (150 PJI cases) were selected randomly. To successfully predict the state of PJI based on MSIS criteria, the NLP technique was applied to all consultation notes, surgical notes, pathology reports, and microbiological reports. After extracting the existence of sinus tract, purulence, pathologic evidence of inflammation, and growth of bacterial isolates from the affected TJA, the algorithm obtained an f1-score between 0.771 and 0.909.

\textbf{D) Medical Language Translation} 

Most researchers, except specialists, have limited knowledge of EHRs because they contain specialised medical terms, acronyms, and a distinct structure and writing style. Translating medical writings into a more understandable form for laypeople is known as medical language translation. For example, the term "peripheral edema" might be substituted with "ankle swelling".  There are only a few research have been conducted on the topic of EHR simplifying. Weng et al. \cite{weng2019unsupervised} used an unsupervised task of text simplification to medical documentation in order to simplify them. Manually annotating text with simplified versions using unsupervised algorithms helps to alleviate the lack of text. They employ skip-gram embeddings learnt from 2 different clinical corpora: MIMIC-III, which has a substantial amount of medical terminology, and MedlinePlus \cite{al2018survey}, which is oriented toward laypeople. These complex and basic phrase embeddings are aligned using a bilingual dictionary induction model, which also initialises a denoising autoencoder. This autoencoder takes as input a sentence written by a doctor, converts it into a simplified translation using a language model, and then reconstructs the original sentence through the translation. On the other hand, a human-annotated medical language translation dataset called MedLane was introduced by Luo et al. \cite{luo2020benchmark}. It aligns professional medical language with expressions that the average person can understand. For training, validation, and testing, it contains 12,801/1,015/1,016 samples, respectively. In addition, they presented the PMBERT-MT model, which employs the pre-trained PubMedBERT \cite{gu2021domain} and carries out translation training using MedLane.

\textbf{E) Medical Disease Prognosis}

Additionally, research in medical imaging is currently in the limelight, which was very useful in the early stages of the COVID-19 outbreak. It should be noted that CNN-based methods were came into the attention to the research community and numerous concepts are currently being developed to solve specific cases. Recently, Bhosale et al. \cite{bhosale2023puldi} introduces a unique CNN model (PulDi-COVID) for the CXI-based detection of nine illnesses (atelectasis, bacterial pneumonia, cardiomegaly, covid19, effusion, infiltration, no-finding, pneumothorax, viral-Pneumonia). Utilizing COVID-19 and CXI data for chronic lung diseases, a variety of transfer-learning models are trained, including VGG16, ResNet50, VGG19, DenseNet201, MobileNetV2, NASNetMobile, ResNet152V2, and DenseNet169. The complete dataset contains a subset of CXI associated with a variety of lung diseases and COVID-19 as well as healthy patients. Furthermore, Bhosale et al. \cite{bhosale2023puldi} select six illnesses from fourteen ChestX-ray8 classifications for the sake of experimentation: atelectasis, cardiomegaly, effusion, infiltration, no-finding/healthy, and pneumothorax. The suggested framework has the greatest achieved accuracy on the dataset utilized in the experiment, with an accuracy of 99.70\%, precision of 98.68\%, recall of 98.67\%, F1-score of 98.67\%, minimal zero-one loss of 12 and error rate of 1.33\%. The proposed model PulDi-COVID has demonstrated superior performance to earlier developed methods. In order to reduce patient severity and mortality, the COVID-19 speedy detection requirements with various lung diseases can be successfully met by the suggested SSE method with PulDi-COVID.

\section{Analysis of the literature}
\label{discussions}

This section will discuss findings based on the retrieved articles.
First, we discuss data types and quantities in \ref{DTQ}. Second, the clinical free text preprocessing pipeline is shown in \ref{text_preprocessing}. The most frequently used ML and DL models are illustrated in \ref{EM}. A comparison of frequently used models is explained in \ref{comparative_analysis}. Model evaluation matrices and commonly used feature extraction methods are illustrated in \ref{evaluation_matrix} and \ref{word_embeddings}. Finally, clinical settings are presented in \ref{CSC}.

\vspace{5mm}
\begin{figure}[H]
  \includegraphics[width=\linewidth]{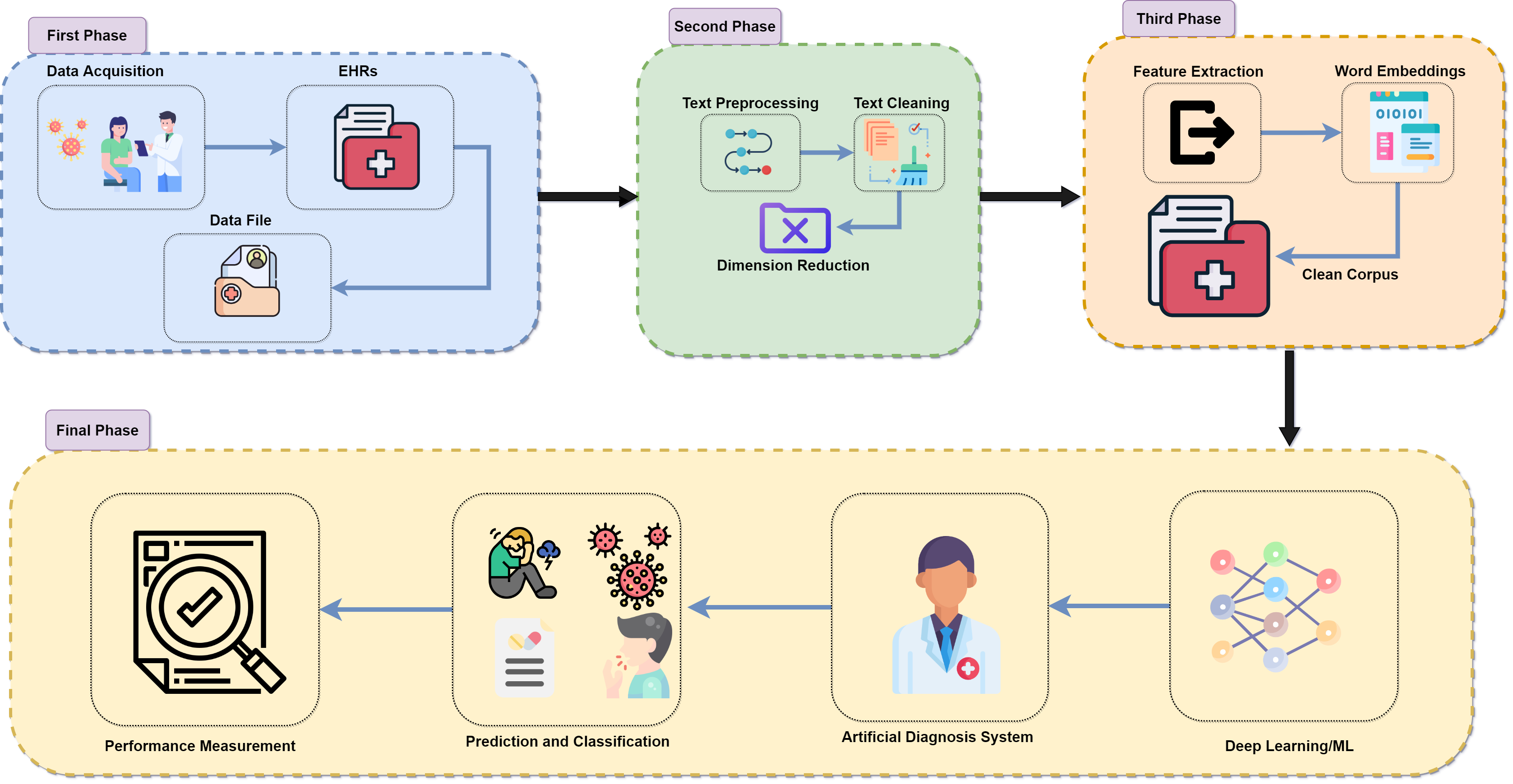}
  \centering
  \captionsetup{justification=centering}
  \caption{Analysing EHRs using DL and ML algorithms.}
  \label{fig:overall_sketch}
\end{figure}

\subsection{Data Type and Quantity}

\label{DTQ}
The Clinical Practice Research Datalink (CPRD) dataset has been used only in one article~\cite{li2020behrt}; all other articles have used electronic health record data. This section describes two essential factors: (a) a comparison of overall insights of data and (b) details of preprocessing pipelines used.

\subsubsection{Comparison of Overall Insights of Data}
Table~\ref{tab:table_data_type} presents studies that reported the following eight parameters: (1) dataset, (2) sample, (3) funding status, (4) universal availability, (5) data type, (6) diagnostic tool, (7) design/settings, and (8) data format. Most of these studies ($n=21$) were well funded, indicating that research on clinical NLP has become essential for improving clinical outcomes in recent years. Table~\ref{tab:table_data_type} groups type of information recorded in EHR, which include clinical notes ($n=14$), clinical narratives ($n=5$), echocardiography reports ($n=2$), medication administration records ($n=4$), lung cancer data ($n=1$), mental health ($n=1$), demographic data ($n=2$) and discharge summary ($n=1$). Due to patients' confidentiality, the availability of clinical data is quite challenging. Most EHR data researched was not open to public access and was not available to access upon study completion.

We also retrieved the parameters of the studies such as diagnostic tool, research design/setting, and data format. These studies can further be clustered into observational studies ($n=8$), experimental studies ($n=24$), and case studies ($n=1$) with a variety of diagnostic tools utilised in both cohort and empirical research. The International Classification of Diseases (ICD) was one of the most widely used approaches to assign codes against a specific disease. The data format of each observational and experimental study was unstructured when obtained from the various EHR sources.

\vspace{5mm}
\begin{table}[H]
\caption{Data type and size with relevant parameters}
\label{tab:table_data_type}
\resizebox{\columnwidth}{!}{%
\begin{tabular}{|c|c|c|c|c|c|c|c|}

\hline
Dataset & Sample & Funded Research & Publicly Available & Data Type & Diagnostic Tool & Design/Setting & Structured \\ \hline
EHRs\cite{lindvall2019natural} & 302 & $\chi$ & $\chi$ & Clinical   Notes & ICD & Cohort & $\chi$ \\ \hline
EHRs\cite{afzal2018natural} & 792 & $\sqrt{ }$ & $\chi$  & Clinical   Notes & ICD & Cohort & $\chi$  \\ \hline
EHRs\cite{dorr2019identifying} & 681 & $\chi$ & $\chi$ & $\chi$ & $\chi$ & $\chi$ & $\chi$ \\ \hline
EHRs\cite{sholle2019underserved} & 16665 & $\sqrt{ }$ & $\chi$ & Clinical Notes & $\chi$ & Experimental   Research & $\chi$ \\ \hline
EHRs\cite{miller2020experiences} & 8000 & $\sqrt{ }$ & $\chi$ & Clinical   Notes & Apache   cTAKES & Experimental   Research & $\chi$ \\ \hline
EHRs\cite{hong2019developing} & 1237 & $\sqrt{ }$ & $\chi$ & Clinical   Discharge Summaries & Phenotyping   Framework. & Case   Study & $\chi$ \\ \hline
Forensic   EHRs\cite{van2018risk} & 6865 & $\chi$ & $\chi$ & Case   Notes & $\chi$ & Experimental   Research & $\chi$ \\ \hline
EHRs\cite{shi2019using} & 291 & $\sqrt{ }$ &  $\chi$ & Clinical   Notes & ACS   NSQIP & Cohort & $\sqrt{ }$ \\ \hline
EHRs\cite{rajendran2020extracting} & 781 & $\sqrt{ }$ & $\chi$  & Clinical   Notes & $\chi$ & Experimental   Research & $\sqrt{ }$ \\ \hline
EHRs\cite{zhao2020incorporating} & 7853 & $\chi$ & $\chi$ & $\chi$ & ICD & Cohort & $\chi$ \\ \hline
EHRs\cite{kogan2020assessing} & 7149 & $\chi$ & $\chi$ & Physician   Notes & $\chi$ & Experimental   Research & $\chi$ \\ \hline
I2B2\cite{gligic2020named} & 4605 & $\sqrt{ }$ & $\sqrt{ }$ & Medication   Administration Record & $\chi$ & Experimental   Research & $\chi$ \\ \hline
EHRs\cite{gligic2020named} & 6861 & $\sqrt{ }$ & $\chi$ & Clinical   Notes & ICD & Cohort & $\chi$ \\ \hline
EHRs\cite{weissler2020use} & $\chi$  & $\chi$  & $\chi$  & Medical   Language & $\chi$  & Experimental   Research & $\chi$  \\ \hline
CRIS\cite{vaci2020natural} & $\chi$  & $\sqrt{ }$ & $\chi$ & Medication   Administration Record & $\chi$ & Experimental   Research & $\chi$ \\ \hline
EHRs\cite{leiter2020deep} & 154 & $\chi$ & $\chi$ & Clinical   Notes & $\chi$ & Experimental   Research & $\chi$ \\ \hline
EHRs\cite{suryanarayanan2020timely} & 700,000 & $\chi$ & $\chi$ & Medication   Administration Record & $\chi$ & Experimental   Research & $\chi$ \\ \hline
EHRs\cite{steinberg2021language} & $\chi$ & $\chi$ & $\chi$ & Patient   Demographic Data & $\chi$ & Experimental   Research & $\chi$ \\ \hline
EHRs\cite{lee2021identifying} & 92151 & $\sqrt{ }$ & $\chi$ & Mental Health  Disorder & $\chi$ & Experimental   Research & $\chi$ \\ \hline
EHRs\cite{yuan2021performance} & 76   643 & $\sqrt{ }$ & $\chi$ & Lung   Cancer & ICD & Cohort & $\chi$ \\ \hline
EHRs\cite{fu2022ascertainment} & 150 & $\sqrt{ }$ &  & Clinical   Notes & $\chi$ & Experimental   Research & $\chi$ \\ \hline
EHRs\cite{solomon2021large} & 1003 & $\sqrt{ }$ & $\chi$  & Echocardiogram   Reports & $\chi$ & Experimental   Research & $\chi$ \\ \hline
EHRs\cite{deng2021natural} & 1052 & $\sqrt{ }$  & $\chi$  & Clinical   Notes & $\chi$  & Experimental   Research & $\chi$  \\ \hline
EHRs\cite{tsui2021natural} & 798   665 &  $\sqrt{ }$ & $\chi$  & Clinical   Notes & ICD & Cohort & $\chi$ \\ \hline
EHRs\cite{bharimalla2021blockchain} & 17   235 & $\chi$ & $\chi$ & Clinical   Narratives & $\chi$ & Experimental   Research & $\chi$ \\ \hline
EHRs\cite{steele2018machine} & 586 &  $\sqrt{ }$ & $\chi$ & Clinical   Narratives & $\chi$ & Experimental   Research & $\chi$ \\ \hline
EHRs\cite{glicksberg2018automated} & $\chi$ & $\chi$ & $\chi$ & Clinical   Narratives & $\chi$ & Experimental   Research & $\chi$ \\ \hline
EHRs\cite{ye2018prediction} & 823,627 & $\chi$ & $\chi$ & Clinical   Narratives & $\chi$ & Experimental   Research & $\chi$ \\ \hline
CPRD\cite{li2020behrt} & 674 &  $\sqrt{ }$ & $\chi$ & Clinical   Narratives & $\chi$ & Experimental   Research & $\chi$ \\ \hline
EHRs\cite{afshar2019subtypes} & 1000 & $\chi$  & $\chi$  & Medication   Administration Record & ICD & Cohort & $\chi$  \\ \hline
EHRs\cite{zheng2017machine} & 300 & $\chi$  & $\chi$  & $\chi$  & $\chi$  & Experimental   Research & $\chi$  \\ \hline
EHRs\cite{wu2020using} & 500 & $\sqrt{ }$ & $\chi$  & Clinical   Notes & $\chi$  & Experimental   Research & $\chi$  \\ \hline
EHRs\cite{jonnalagadda2017text} & 198 & $\sqrt{ }$ & $\chi$  & Echocardiography   Reports & ICD & Cohort & $\chi$ \\ \hline
EHRs\cite{wu2020using} & 820 & $\sqrt{ }$ & $\chi$ & Patient   Demographics & $\chi$ & Experimental   Research & $\sqrt{ }$ \\ \hline
EHRs\cite{downs2017detection} & $\chi$ & $\sqrt{ }$ & $\chi$ & Clinical   Narratives & $\chi$ & Experimental   Research & $\chi$ \\ \hline
\end{tabular}%
}
\end{table}

\subsubsection{Clinical Free Text Preprocessing Pipeline}
\label{text_preprocessing}
Understanding what techniques researchers frequently use in clinical text processing is essential. Finding the right direction to process clinical free text is vital to understanding the free text processing settings. Table~\ref{tab:data_processing} presents data preprocessing methods used in the reviewed articles. We compare commonly used data preprocessing techniques such as commercial, manual, electronic, and distributed. Our analysis reveals that the researchers did not explain any details of the clinical text preprocessing settings in the articles listed in Table~\ref{tab:data_processing}. Although structured data was used in two manuscripts \cite{shi2019using} \cite{rajendran2020extracting}, we did not find comprehensive approaches to clinical text preprocessing.

\vspace{5mm}
\begin{table}[H]
\caption{Comparison of data preprocessing methods}
\label{tab:data_processing}
\resizebox{\columnwidth}{!}{%
\begin{tabular}{|c|c|c|c|c|c|c|}

\hline
Dataset & Data Type & Structured & Commercial Data Processing & Manual Processing & Electronic Processing & Distributed Processing \\ \hline
EHRs\cite{lindvall2019natural} & Clinical   Notes & $\chi$  & $\chi$  & $\chi$  & $\chi$  & $\chi$  \\ \hline
EHRs\cite{afzal2018natural} & Clinical   Notes & $\chi$  & $\chi$  & $\chi$  & $\chi$  & $\chi$  \\ \hline
EHRs\cite{dorr2019identifying}  & $\chi$  & $\chi$  & $\chi$  & $\chi$  & $\chi$  & $\chi$  \\ \hline
EHRs\cite{sholle2019underserved} & Clinical   Notes & $\chi$  & $\chi$  & $\chi$  & $\chi$  & $\chi$  \\ \hline
EHRs\cite{miller2020experiences} & Clinical   Notes & $\chi$  & $\chi$  & $\chi$  & $\chi$  & $\chi$  \\ \hline
EHRs\cite{hong2019developing} & Clinical   Discharge Summaries & $\chi$  & $\chi$  & $\chi$  & $\chi$  & $\chi$  \\ \hline
Forensic   EHRs\cite{van2018risk} & Case   Notes & $\chi$  & $\chi$  & $\chi$  & $\chi$  & $\chi$  \\ \hline
EHRs\cite{shi2019using} & Clinical   Notes & $\sqrt{ }$ & $\chi$  & $\chi$  & $\chi$  & $\chi$  \\ \hline
EHRs\cite{rajendran2020extracting}  & Clinical   Notes & $\sqrt{ }$ & $\chi$ & $\chi$ & $\chi$ & $\chi$ \\ \hline
EHRs\cite{zhao2020incorporating}  & $\chi$  & $\chi$  & $\chi$  & $\chi$  & $\chi$  & $\chi$  \\ \hline
EHRs\cite{kogan2020assessing} & Physician   Notes & $\chi$  & $\chi$  & $\chi$  & $\chi$  & $\chi$  \\ \hline
EHRs\cite{gligic2020named} & Clinical   Notes & $\chi$  & $\chi$  & $\chi$  & $\chi$  & $\chi$  \\ \hline
EHRs\cite{weissler2020use}  & Medical   Language & $\chi$  & $\chi$  & $\chi$  & $\chi$  & $\chi$  \\ \hline
CRIS\cite{vaci2020natural} & Medication   Administration Record & $\chi$  & $\chi$  & $\chi$  & $\chi$  & $\chi$  \\ \hline
EHRs\cite{leiter2020deep} & Clinical   Notes & $\chi$  & $\chi$  & $\chi$  & $\chi$  & $\chi$  \\ \hline
EHRs\cite{suryanarayanan2020timely} & Medication   Administration Record & $\chi$  & $\chi$  & $\chi$  & $\chi$  & $\chi$  \\ \hline
EHRs\cite{steinberg2021language} & Patient   Demographic Data & $\chi$  & $\chi$  & $\chi$  & $\chi$  & $\chi$  \\ \hline
EHRs\cite{lee2021identifying} & Clinical   Notes & $\chi$  & $\chi$  & $\chi$  & $\chi$  & $\chi$  \\ \hline
EHRs\cite{yuan2021performance} & Lung   Cancer & $\chi$  & $\chi$  & $\chi$  & $\chi$  & $\chi$  \\ \hline
EHRs\cite{fu2022ascertainment} & Clinical   Notes & $\chi$  & $\chi$  & $\chi$  & $\chi$  & $\chi$  \\ \hline
EHRs\cite{solomon2021large} & Physician-adjudicated   echocardiogram reports & $\chi$  & $\chi$  & $\chi$  & $\chi$  & $\chi$  \\ \hline
EHRs\cite{deng2021natural} & Clinical   Notes & $\chi$  & $\chi$  & $\chi$  & $\chi$  & $\chi$  \\ \hline
EHRs\cite{tsui2021natural} & Clinical   Notes & $\chi$  & $\chi$  & $\chi$  & $\chi$  & $\chi$  \\ \hline
EHRs\cite{bharimalla2021blockchain}  & Clinical   Narratives & $\chi$  & $\chi$  & $\chi$  & $\chi$  & $\chi$  \\ \hline
EHRs\cite{steele2018machine} & Clinical   Narratives & $\chi$  & $\chi$  & $\chi$  & $\chi$  & $\chi$  \\ \hline
EHRs\cite{glicksberg2018automated} & Clinical   Narratives & $\chi$  & $\chi$  & $\chi$  & $\chi$  & $\chi$  \\ \hline
EHRs\cite{ye2018prediction}  & Clinical   Narratives & $\chi$  & $\chi$  & $\chi$  & $\chi$  & $\chi$  \\ \hline
CPRD\cite{li2020behrt} & Clinical   Narratives & $\chi$  & $\chi$  & $\chi$  & $\chi$  & $\chi$  \\ \hline
EHRs\cite{afshar2019subtypes} & Medication   Administration Record & $\chi$  & $\chi$  & $\chi$  & $\chi$  & $\chi$  \\ \hline
EHRs\cite{zheng2017machine} & $\chi$  & $\chi$  & $\chi$  & $\chi$  & $\chi$  & $\chi$  \\ \hline

EHRs\cite{downs2017detection} & Echocardiography   Reports & $\chi$  & $\chi$  & $\chi$  & $\chi$  & $\chi$  \\ \hline
EHRs\cite{wu2020using} & Patient   Demographics & $\chi$  & $\chi$  & $\chi$  & $\chi$  & $\chi$  \\ \hline
EHRs\cite{downs2017detection} & Clinical   Narratives & $\chi$  & $\chi$  & $\chi$  & $\chi$  & $\chi$  \\ \hline
\end{tabular}%
}
\end{table}

\subsection{Models}
\label{EM}

\subsubsection{Frequently Used ML Models}

Experimental techniques that were effective when implemented on EHRs were Logistic Regression (LR), Support Vector Machine (SVM), eXtreme Gradient Boosting (XGBoost), AdaBoost, Random Forest (RF), Linear Regression (LR), Naïve Bayes (NB), Gradient Boosting (GB), and Decision Tree (DT) models. These ML and DL-based algorithms were applied to conduct various NLP tasks, including classification, prediction, word embedding, text summarisation, language modelling, ICD-10 classification, clinical notes analysis, mental health issue identification and medical dialogue analysis. The two most prominent NLP tasks in recent years have been classification ($n=15$) and prediction ($n=14$).

\begin{figure}[H]
    \centering
    \captionsetup{justification=centering}
    \includegraphics[width=0.7\textwidth]{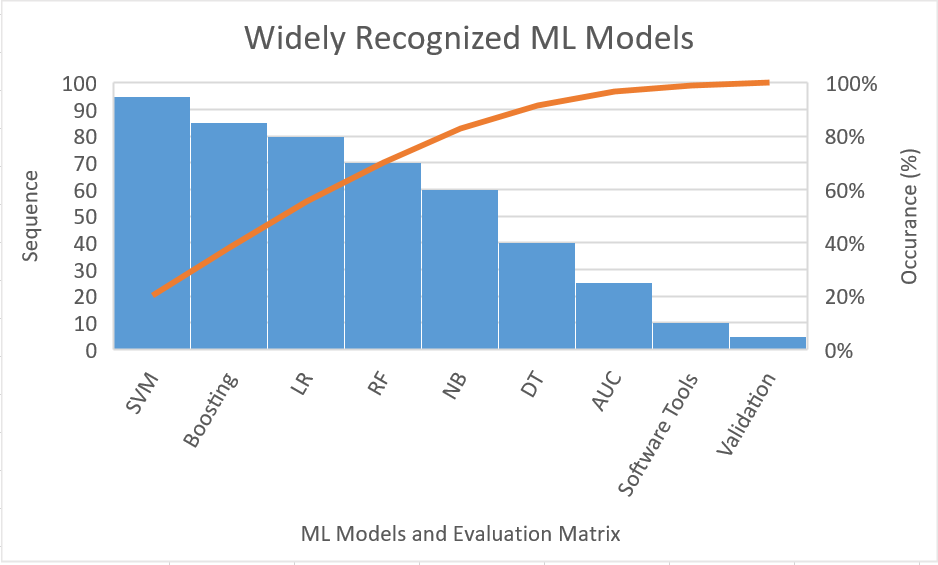}
    \caption{Visual presentation of recognised machine learning models and mapping their cumulative frequencies.}
    
    \label{fig:ML_MODEL}
    
\end{figure}

The bar graph in Figure ~\ref{fig:ML_MODEL} compares the widely accepted ML models for medical NLP used for EHRs. In short, Support Vector Machine (SVM) and boosting algorithms have been the most widely utilised models applied to electronic health record data for many years. 
Now turning back to the details, Figure ~\ref{fig:ML_MODEL} clearly explain that the use of the SVM model for clinical free text analysis has increased rapidly by 95\% in recent years, showing that scholars have concentrated more on utilizing this approach in recent years. On the other hand, the use of the Decision Tree (DT) model was the lowest among other classical ML algorithms at 40\%. It is obvious that boosting strategies such as AdaBoost and XGBoost were used significantly in selected articles. It is also noticeable that, in recent years, about four-fifths of the Logistic Regression (LR) model has been applied to analyse medical free text, as can be clearly seen from Figure ~\ref{fig:ML_MODEL}. Finally, model evaluation indicators and automated software tools were used in a very small number of articles.

\subsubsection{Frequently Used DL Models}

In addition to ML models, various DL-based strategies from data concerning health are also applied to make automated decisions. Table~\ref{tab:top_models} explains the frequently used DL models and compares them with the evaluation metrics techniques. Moreover, we were unable to determine why evaluation metrics were not discussed in many research articles. One possible reason is that the model's performance was rather satisfactory; therefore, evaluation via metrics was not used.

\begin{figure}[H]
    \centering
    \captionsetup{justification=centering}
    \includegraphics[width=0.7\textwidth]{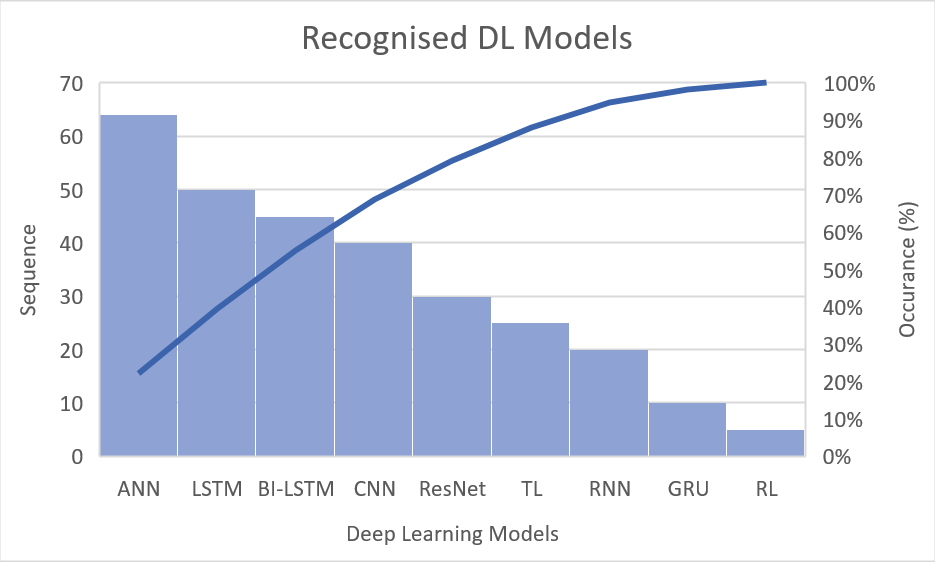}
    \caption{Deep learning models and their cumulative frequency mapping. }
    
    \label{fig:my_label_CFM}
\end{figure}

The bar chart in Figure~\ref{fig:my_label_CFM} illustrates the recognised DL models and their cumulative frequency mapping. Overall, it can be seen that Neural Network (NN) was the most frequently used model applied to the electronic health records for analysing clinical free text. Artificial neural networks (ANN), commonly referred to as neural networks or neural nets, are inspired by biological brain networks. An ANN is comprised of a network of interconnected units or nodes known as artificial neurons, which loosely resemble the neurons of a biological brain.


The other common models identified in the included studies were Long Short-Term Memory (LSTM), Bidirectional Long Short-Term Memory (BI-LSTM), Convolutional Neural Network (CNN), Residual Neural Network (ResNet), Transfer Learning (TL), Recurrent Neural Network (RNN), Gated Recurrent Units (GRU) and Representation Learning (RL).

\vspace{5mm}
\begin{table}[H]
\caption{Top frequently utilised traditional models with evaluation metrics and validation techniques of the existing research articles.}
\label{tab:top_models}
\resizebox{\columnwidth}{!}{%
\begin{tabular}{|c|ccccc|c|}
\hline
\multirow{2}{*}{Model used} & \multicolumn{5}{c|}{Evaluation metrics} & \multirow{2}{*}{Validation} \\ \cline{2-6} &
 \multicolumn{1}{c|}{AUC} & \multicolumn{1}{c|}{Accuracy} & \multicolumn{1}{c|}{P} & \multicolumn{1}{c|}{R} & \multicolumn{1}{c|}{F1 Score}  \\ \hline
 
Elastic Net~\cite{dorr2019identifying} & 
\multicolumn{1}{c|}{$\chi$} & \multicolumn{1}{c|}{$\chi$} & \multicolumn{1}{c|}{$\chi$} & \multicolumn{1}{c|}{$\chi$} & $\chi$ & \multicolumn{1}{c|}{$\chi$} \\ \hline

NLP, EPA~\cite{sholle2019underserved} & 
\multicolumn{1}{c|}{0.98} & \multicolumn{1}{c|}{$\chi$} & \multicolumn{1}{c|}{$\chi$} & \multicolumn{1}{c|}{$\chi$} & $\chi$ & \multicolumn{1}{c|}{$\chi$} \\ \hline

XGBoost~\cite{miller2020experiences} & 
\multicolumn{1}{c|}{0.91} & \multicolumn{1}{c|}{$\chi$} & \multicolumn{1}{c|}{$\chi$} & \multicolumn{1}{c|}{$\chi$} & $\chi$ & \multicolumn{1}{c|}{$\chi$} \\ \hline

BEHRT~\cite{hong2019developing} & 
\multicolumn{1}{c|}{0.91} & \multicolumn{1}{c|}{$\chi$} & \multicolumn{1}{c|}{$\sqrt{ }$} & \multicolumn{1}{c|}{$\chi$} & $\chi$ & \multicolumn{1}{c|}{$\chi$} \\ \hline

NLP, LCA~\cite{van2018risk} & 
\multicolumn{1}{c|}{$\chi$} & \multicolumn{1}{c|}{$\chi$} & \multicolumn{1}{c|}{$\chi$} & \multicolumn{1}{c|}{$\chi$} & $\chi$ & \multicolumn{1}{c|}{$\chi$} \\ \hline

K-NN, SVM, NB, RF~\cite{shi2019using} & 
\multicolumn{1}{c|}{0.98} & \multicolumn{1}{c|}{$\chi$} & \multicolumn{1}{c|}{$\chi$} & \multicolumn{1}{c|}{$\chi$} & $\chi$ & \multicolumn{1}{c|}{$\chi$} \\ \hline

UMLS, MedLEE, NV~\cite{rajendran2020extracting} & 
\multicolumn{1}{c|}{0.9} & \multicolumn{1}{c|}{$\chi$} & \multicolumn{1}{c|}{$\chi$} & \multicolumn{1}{c|}{$\chi$} & $\chi$ & \multicolumn{1}{c|}{$\chi$} \\ \hline

RF~\cite{zhao2020incorporating} & 
\multicolumn{1}{c|}{0.92} & \multicolumn{1}{c|}{$\chi$} & \multicolumn{1}{c|}{$\chi$} & \multicolumn{1}{c|}{$\chi$} & $\chi$ & \multicolumn{1}{c|}{$\chi$} \\ \hline

N/A~\cite{kogan2020assessing} & 
\multicolumn{1}{c|}{$\chi$} & \multicolumn{1}{c|}{$\chi$} & \multicolumn{1}{c|}{$\chi$} & \multicolumn{1}{c|}{$\sqrt{ }$} & $\chi$ & \multicolumn{1}{c|}{$\chi$} \\ \hline



N/A~\cite{gavrilov2020feature} & 
\multicolumn{1}{c|}{$\chi$} & \multicolumn{1}{c|}{97} & \multicolumn{1}{c|}{$\chi$} & \multicolumn{1}{c|}{$\chi$} & $\chi$ & \multicolumn{1}{c|}{$\chi$} \\ \hline

N/A~\cite{vaci2020natural} & 
\multicolumn{1}{c|}{$\chi$} & \multicolumn{1}{c|}{86.81} & \multicolumn{1}{c|}{$\chi$} & \multicolumn{1}{c|}{$\chi$} & $\chi$ & \multicolumn{1}{c|}{$\chi$} \\ \hline

N/A~\cite{leiter2020deep} & 
\multicolumn{1}{c|}{$\chi$} & \multicolumn{1}{c|}{$\chi$} & \multicolumn{1}{c|}{$\chi$} & \multicolumn{1}{c|}{85} & {$\sqrt{ }$} & \multicolumn{1}{c|}{$\chi$} \\ \hline

N/A~\cite{zhu2020using} & 
\multicolumn{1}{c|}{$\chi$} & \multicolumn{1}{c|}{93.8} & \multicolumn{1}{c|}{$\chi$} & \multicolumn{1}{c|}{$\chi$} & $\chi$ & \multicolumn{1}{c|}{$\chi$} \\ \hline

N/A~\cite{weiner2018semi} & 
\multicolumn{1}{c|}{$\chi$} & \multicolumn{1}{c|}{$\chi$} & \multicolumn{1}{c|}{$\chi$} & \multicolumn{1}{c|}{$\sqrt{ }$} & $\chi$ & \multicolumn{1}{c|}{$\chi$} \\ \hline

CLM BR ~\cite{steinberg2021language} & 
\multicolumn{1}{c|}{$\chi$} & \multicolumn{1}{c|}{$\chi$} & \multicolumn{1}{c|}{$\chi$} & \multicolumn{1}{c|}{$\chi$} & $\chi$ & \multicolumn{1}{c|}{$\chi$} \\ \hline

N/A~\cite{sivarethinamohan2021envisioning} & 
\multicolumn{1}{c|}{$\chi$} & \multicolumn{1}{c|}{$\chi$} & \multicolumn{1}{c|}{$\chi$} & \multicolumn{1}{c|}{$\chi$} & $\chi$ & \multicolumn{1}{c|}{$\chi$} \\ \hline

N/A~\cite{dymek2021building} & 
\multicolumn{1}{c|}{$\chi$} & \multicolumn{1}{c|}{$\chi$} & \multicolumn{1}{c|}{$\sqrt{ }$} & \multicolumn{1}{c|}{$\chi$} & $\chi$ & \multicolumn{1}{c|}{$\chi$} \\ \hline

NLP and supervised ML\cite{lee2021identifying} & 
\multicolumn{1}{c|}{93.3} & \multicolumn{1}{c|}{$\chi$} & \multicolumn{1}{c|}{$\chi$} & \multicolumn{1}{c|}{$\chi$} & $\chi$ & \multicolumn{1}{c|}{$\chi$} \\ \hline

LASSO~\cite{shen2021analysis} & 
\multicolumn{1}{c|}{0.58} & \multicolumn{1}{c|}{$\chi$} & \multicolumn{1}{c|}{$\chi$} & \multicolumn{1}{c|}{$\chi$} & $\chi$ & \multicolumn{1}{c|}{$\chi$} \\ \hline

Sentiment Analysis, Cognition Engine and NLP techniques\cite{levis2021natural} & 
\multicolumn{1}{c|}{0.58} & \multicolumn{1}{c|}{N/A} & \multicolumn{1}{c|}{$\chi$} & \multicolumn{1}{c|}{$\chi$} & $\chi$ & \multicolumn{1}{c|}{$\chi$} \\ \hline

PheCAP~\cite{irving2021using} & 
\multicolumn{1}{c|}{$\chi$} & \multicolumn{1}{c|}{$\sqrt{ }$} & \multicolumn{1}{c|}{$\chi$} & \multicolumn{1}{c|}{$\chi$} & $\chi$ & \multicolumn{1}{c|}{$\chi$} \\ \hline

VHA~\cite{yuan2021performance} & 
\multicolumn{1}{c|}{$\chi$} & \multicolumn{1}{c|}{94.4} & \multicolumn{1}{c|}{$\chi$} & \multicolumn{1}{c|}{$\chi$} & $\chi$ & \multicolumn{1}{c|}{$\chi$} \\ \hline


N/A~\cite{fu2022ascertainment} & 
\multicolumn{1}{c|}{$\chi$} & \multicolumn{1}{c|}{$\sqrt{ }$} & \multicolumn{1}{c|}{$\chi$} & \multicolumn{1}{c|}{$\chi$} & $\chi$ & \multicolumn{1}{c|}{$\chi$} \\ \hline

Topic modeling+LDA~\cite{viani2021natural} & 
\multicolumn{1}{c|}{$\chi$} & \multicolumn{1}{c|}{81} & \multicolumn{1}{c|}{$\chi$} & \multicolumn{1}{c|}{$\chi$} & $\chi$ & \multicolumn{1}{c|}{$\chi$} \\ \hline

cTAKES~\cite{tedeschi2021classifying} & 
\multicolumn{1}{c|}{$\chi$} & \multicolumn{1}{c|}{96.7} & \multicolumn{1}{c|}{$\chi$} & \multicolumn{1}{c|}{$\chi$} & $\chi$ & \multicolumn{1}{c|}{$\chi$} \\ \hline

N/A~\cite{moore2021ascertaining} & 
\multicolumn{1}{c|}{$\chi$} & \multicolumn{1}{c|}{98.2} & \multicolumn{1}{c|}{$\chi$} & \multicolumn{1}{c|}{$\chi$} & $\chi$ & \multicolumn{1}{c|}{$\chi$} \\ \hline

LR, RF, SVM~\cite{fu2021automated} & 
\multicolumn{1}{c|}{$\chi$} & \multicolumn{1}{c|}{71} & \multicolumn{1}{c|}{$\chi$} & \multicolumn{1}{c|}{$\chi$} & $\chi$ & \multicolumn{1}{c|}{$\chi$} \\ \hline

Sag, Meta map, SHAP~\cite{viani2021natural} & 
\multicolumn{1}{c|}{$\chi$} & \multicolumn{1}{c|}{96} & \multicolumn{1}{c|}{$\chi$} & \multicolumn{1}{c|}{$\chi$} & $\chi$ & \multicolumn{1}{c|}{$\chi$} \\ \hline

CNN-LSTM, ResNet-LSTM~\cite{deng2021natural} & 
\multicolumn{1}{c|}{$\chi$} & \multicolumn{1}{c|}{99} & \multicolumn{1}{c|}{$\chi$} & \multicolumn{1}{c|}{$\chi$} & $\chi$ & \multicolumn{1}{c|}{$\chi$} \\ \hline

LF, Sense2vec, OxCRIS~\cite{deng2021natural} & 
\multicolumn{1}{c|}{$\chi$} & \multicolumn{1}{c|}{95.7} & \multicolumn{1}{c|}{$\chi$} & \multicolumn{1}{c|}{$\chi$} & $\chi$ & \multicolumn{1}{c|}{$\chi$} \\ \hline

NLP, DL~\cite{kormilitzin2021med7} & 
\multicolumn{1}{c|}{$\chi$} & \multicolumn{1}{c|}{$\chi$} & \multicolumn{1}{c|}{$\chi$} & \multicolumn{1}{c|}{$\chi$} & $\chi$ & \multicolumn{1}{c|}{$\chi$} \\ \hline

RF, LASSO, EXGB, NV~\cite{jain2021nlp} & 
\multicolumn{1}{c|}{0.95} & \multicolumn{1}{c|}{$\chi$} & \multicolumn{1}{c|}{$\chi$} & \multicolumn{1}{c|}{$\chi$} & $\chi$ & \multicolumn{1}{c|}{$\chi$} \\ \hline

N/A~\cite{weng2017medical} & 
\multicolumn{1}{c|}{$\chi$} & \multicolumn{1}{c|}{$\chi$} & \multicolumn{1}{c|}{97} & \multicolumn{1}{c|}{$\chi$} & $\chi$ & \multicolumn{1}{c|}{$\chi$} \\ \hline

ASUDS, LRM~\cite{ayre2021developing} & 
\multicolumn{1}{c|}{$\chi$} & \multicolumn{1}{c|}{94} & \multicolumn{1}{c|}{$\chi$} & \multicolumn{1}{c|}{$\chi$} & $\chi$ & \multicolumn{1}{c|}{$\chi$} \\ \hline

Cohort study~\cite{ni2021automated} & 
\multicolumn{1}{c|}{$\chi$} & \multicolumn{1}{c|}{$\chi$} & \multicolumn{1}{c|}{$\sqrt{ }$} & \multicolumn{1}{c|}{$\sqrt{ }$} & $\chi$ & \multicolumn{1}{c|}{$\chi$} \\ \hline

ICD9-CM, CPT and NLP techniques\cite{lindvall2019natural} & 
\multicolumn{1}{c|}{$\chi$} & \multicolumn{1}{c|}{$\chi$} & \multicolumn{1}{c|}{$\chi$} & \multicolumn{1}{c|}{85.7–92.9} & {$\sqrt{ }$}& \multicolumn{1}{c|}{$\chi$} \\ \hline

Expert-driven Queries+NLP~\cite{afzal2018natural} & 
\multicolumn{1}{c|}{$\chi$} & \multicolumn{1}{c|}{$\chi$} & \multicolumn{1}{c|}{$\chi$} & \multicolumn{1}{c|}{$\chi$} & $\chi$ & \multicolumn{1}{c|}{$\sqrt{ }$} \\ \hline

Rule-based NLP~\cite{dorr2019identifying} & 
\multicolumn{1}{c|}{$\chi$} & \multicolumn{1}{c|}{$\chi$} & \multicolumn{1}{c|}{$\chi$} & \multicolumn{1}{c|}{$\chi$} & $\chi$ & \multicolumn{1}{c|}{$\sqrt{ }$} \\ \hline

cTAKES NLP Software~\cite{sholle2019underserved} & 
\multicolumn{1}{c|}{$\chi$} & \multicolumn{1}{c|}{$\chi$} & \multicolumn{1}{c|}{$\chi$} & \multicolumn{1}{c|}{$\chi$} & $\chi$ & \multicolumn{1}{c|}{$\chi$} \\ \hline

LR, SVM, DT and RF~\cite{miller2020experiences} & 
\multicolumn{1}{c|}{$\sqrt{ }$} & \multicolumn{1}{c|}{$\sqrt{ }$} & \multicolumn{1}{c|}{$\sqrt{ }$} & \multicolumn{1}{c|}{$\sqrt{ }$} & \multicolumn{1}{c|}{$\sqrt{ }$} & \multicolumn{1}{c|}{$\sqrt{ }$} \\ \hline

LMT, LR, Linear Regression and SVM~\cite{miller2020experiences} & 
\multicolumn{1}{c|}{$\chi$} & \multicolumn{1}{c|}{$\chi$} & \multicolumn{1}{c|}{$\sqrt{ }$} & \multicolumn{1}{c|}{$\sqrt{ }$} & \multicolumn{1}{c|}{$\sqrt{ }$} & \multicolumn{1}{c|}{$\chi$}\\ \hline

Automated Clinical Follow-up Tool~\cite{van2018risk} & 
\multicolumn{1}{c|}{$\sqrt{ }$} & \multicolumn{1}{c|}{$\sqrt{ }$} & \multicolumn{1}{c|}{$\sqrt{ }$} & \multicolumn{1}{c|}{$\sqrt{ }$} & \multicolumn{1}{c|}{$\sqrt{ }$} & \multicolumn{1}{c|}{$\sqrt{ }$} \\ \hline

Regression, SVMs, DT, RF~\cite{kovacs2017correlate} & 
\multicolumn{1}{c|}{$\sqrt{ }$} & \multicolumn{1}{c|}{$\chi$} & \multicolumn{1}{c|}{$\chi$} & \multicolumn{1}{c|}{$\chi$} & $\chi$ & \multicolumn{1}{c|}{$\chi$} \\ \hline

RF, SVM, LR~\cite{zeng2018natural} & 
\multicolumn{1}{c|}{$\chi$} & \multicolumn{1}{c|}{$\chi$} & \multicolumn{1}{c|}{$\sqrt{ }$} & \multicolumn{1}{c|}{$\sqrt{ }$} & \multicolumn{1}{c|}{$\sqrt{ }$} & \multicolumn{1}{c|}{$\chi$}\\ \hline

Supervised and Unsupervised Model~\cite{shi2019using} & 
\multicolumn{1}{c|}{$\sqrt{ }$} & \multicolumn{1}{c|}{$\chi$} & \multicolumn{1}{c|}{$\sqrt{ }$} & \multicolumn{1}{c|}{$\sqrt{ }$} & \multicolumn{1}{c|}{$\sqrt{ }$} & \multicolumn{1}{c|}{$\chi$}\\ \hline

RF, Gradient Boosting, Neural Network, and Linear Regression~\cite{shi2019using} & 
\multicolumn{1}{c|}{$\sqrt{ }$} & \multicolumn{1}{c|}{$\chi$} & \multicolumn{1}{c|}{$\chi$} & \multicolumn{1}{c|}{$\chi$} & \multicolumn{1}{c|}{$\chi$} & \multicolumn{1}{c|}{$\sqrt{ }$}\\ \hline

Transfer Learning and Neural Networks~\cite{gligic2020named} & 
\multicolumn{1}{c|}{$\chi$}  & \multicolumn{1}{c|}{$\chi$} & \multicolumn{1}{c|}{$\chi$} & \multicolumn{1}{c|}{$\chi$} & \multicolumn{1}{c|}{82.4} & \multicolumn{1}{c|}{$\chi$} \\ \hline

PAD-ML and  LASSO approach ~\cite{weissler2020use} & 
\multicolumn{1}{c|}{0.801 and 0.888}  & \multicolumn{1}{c|}{70}& \multicolumn{1}{c|}{90} & \multicolumn{1}{c|}{$\sqrt{ }$} & \multicolumn{1}{c|}{$\chi$} & \multicolumn{1}{c|}{$\sqrt{ }$} \\ \hline

\end{tabular}%
}
\end{table}

\subsection{Comparison of frequently utilised models}

In this sub-section, we compare commonly used ML and DL-based models and discuss their advantages and disadvantages in general. It will provide readers with an understanding of the core information of each model in a clinical free text context.
\label{comparative_analysis}

\vspace{5mm}

\begin{table}[H]
\caption{Comparison of popular ML and DL models applied in the EHRs}
\label{tab:my-table}
\resizebox{\columnwidth}{!}{%
\begin{tabular}{|c|l|l|}
\hline
Model &
  \multicolumn{1}{c|}{Advantage} &
  \multicolumn{1}{c|}{Disadvantage} \\ \hline
SVM &
  \begin{tabular}[c]{@{}l@{}}(a) SVM is very efficient for high dimensional   space. \\    \\ (b) This algorithm uses relatively less memory.\end{tabular} &
  \begin{tabular}[c]{@{}l@{}}(a) Support vector machine is not very efficient for very large datasets.\\ (b) The model is easily affected if the dataset contains overlapping classes and noise.\end{tabular} \\ \hline
Gradient Boosting &
  \begin{tabular}[c]{@{}l@{}}(a) Does not require data scaling and can handle missing values.\\    \\ (b) The algorithm is relatively flexible due to loss function \\ optimisation and hyperparameter tuning.\end{tabular} &
  \begin{tabular}[c]{@{}l@{}}(a) Outliers may overfit the model.\\ (b) Comparatively more time-consuming /slower and requires more memory.\end{tabular} \\ \hline
XGBoost &
  \begin{tabular}[c]{@{}l@{}}(a) Can build decision trees in parallel.\\    \\ (b) Can use distributed computing method for complex models.\end{tabular} &
  (a) XGBoost does not perform so well on sparse and unstructured data. \\ \hline
Logistic Regression &
  \begin{tabular}[c]{@{}l@{}}(a) Logistic regression is easy to implement and interpret. \\ At the same time, it uses relatively less computational resources.\\ (b) Logistic regression works well when the data   are linearly separable.\end{tabular} &
  \begin{tabular}[c]{@{}l@{}}(a) The overfitting tendency of logistic regression is generally low, \\ but the model may overfit if the dataset becomes too dimensional, \\ in which case, dimension reduction should be done before modeling~\cite{sivapalan2011compressive}. \\ \\ (b) If the number of observations becomes less than the number of features, \\ the model will not be valid, and the problem of overfitting will arise.\end{tabular} \\ \hline
Random Forest &
  \begin{tabular}[c]{@{}l@{}}(a) Following the random forest bagging method reduces the probability of \\ being influenced by outliers.\\ \\ (b) It works well for both categorical and continuous data,\end{tabular} &
  \begin{tabular}[c]{@{}l@{}}(a) Complex models require more computational resources when the number \\ of learners is large.\end{tabular} \\ \hline
Naïve Bayes &
  \begin{tabular}[c]{@{}l@{}}(a) It is easy to implement and relatively fast.\\ (b) It also works well for small datasets.\end{tabular} &
  (a) Gives slightly lower accuracy than other algorithms. \\ \hline
Decision Tree &
  \begin{tabular}[c]{@{}l@{}}(a) Decision trees do not require the dataset to be scaled.\\ (b) Decision tree can be explained very easily.\end{tabular} &
  \begin{tabular}[c]{@{}l@{}}(a) Decision trees are not very effective for continuous value prediction in many cases.\\ \\ (b) Decision tree model takes comparatively more time in training.\end{tabular} \\ \hline
Neural Network &
  (a) Multitasking is a common advantage of neural   networks. &
  \begin{tabular}[c]{@{}l@{}}(a) Black Box Nature\\ (b) Hardware dependent\end{tabular} \\ \hline
LSTM &
  (a) The complexity of updating each weight is reduced to O(1). &
  (a) Dropout is much harder to implement in LSTMs. \\ \hline
BI-LSTM &
  (a) Enable   additional training by traversing the input data twice &
  (a) Since BiLSTM has double LSTM cells, so it is costly \\ \hline
CNN &
  (a) Without any human oversight, it automatically discovers significant features. &
  (a) Large training data required \\ \hline
ResNet &
  (a) Large number of layers can be trained easily without increasing the training error percentage. &
  (a) Deeper network usually requires weeks of training. \\ \hline
Transfer Learning (TL) &
  (a) Overcome cost- and time-consuming issues. &
  (a) Problem of negative transfer, i.e., utilizing source domain data/knowledge reduces unfavourably learning performance in the target domain.\\ \hline
Recurrent Neural Network (RNN) &
  (a) When processing temporal, sequential data, like text or videos, RNNs perform better. &
  (a) Gradient vanishing and exploding problems. \\ \hline
\end{tabular}%
}
\end{table}

\subsection{Model Evaluation Metrics}

\label{evaluation_matrix}

Model evaluation metrics take a key role in evaluating the accuracy and performance of a trained model. Our analyses reveal that researchers focused primarily on AUC, Accuracy, Precision, Recall and F1-score among the articles we reviewed. It is noteworthy that the AUC tends to differentiate between the classes of a dataset. The higher the AUC, the better the performance of a model that distinguishes between positive and negative classes. Furthermore, the Confusion matrix measures the precision of all classification techniques. The Confusion Matrix has four distinct values: True Positive (TP), False Positive (FP), True Negative (TN) and False Negative (FN). False Positive of Confusion Matrix is called Type 1 Error, and False Negative is called Type 2 Error. Several approaches are used to evaluate a model's accuracy. For example, TP, TN, FP, and FN are the main determinants of the model's performance.

The following equations (1), (2), (3) and (4) are primarily applied to determine the precision, recall, and f1-score \cite{yacouby2020probabilistic}.

\begin{equation}
    \text { Precision }=\frac{T P}{T P+F P}
\end{equation}

\begin{equation}
    \text { Recall }=\frac{T P}{T P+F N}                 
\end{equation}

\begin{equation}
    F 1=2 \cdot \frac{\text { Precision.Recall }}{\text { Precision }+\text { Recall }} 
\end{equation}

\begin{equation}
    \text { Accuracy }=\frac{T P+T N}{T P+T N+F P+F N}
\end{equation}

\subsection{Word Embedding/Feature Extraction Methods}

\label{word_embeddings}
Table~\ref{tab:my-table_feature_extraction} explains feature extraction approaches used with EHR. From the studies we reviewed, various feature extraction techniques were adopted, most of which are traditional approaches. Feature extraction methods of medical narratives such as word weighting, word embedding, and some open-source tools were considered in various studies. Examples include TF-IDF, BOW, Word2vec, Glove, and FastText. The TF-IDF and BOW methods were incorporated as a word weighting technique, while the Word2vec and Glove were favoured as a word embedding approach. Among the selected papers, TF-IDF ($n=7$), BOW ($n=5$), and Glove ($n=5$) were more frequently utilised than the other methods, such as FastText or BERT transformer models.

\begin{table}[H]
\caption{Feature extraction approaches used with EHRs}
\label{tab:my-table_feature_extraction}
\resizebox{\columnwidth}{!}{%
\begin{tabular}{|c|ccc|ccc|c|cc|}
\hline
\multirow{2}{*}{Context} & \multicolumn{3}{c|}{Word Weighting} & \multicolumn{3}{c|}{Word Embedding} & \multirow{2}{*}{Transformer Approach} & \multicolumn{2}{c|}{Automated Tools} \\ \cline{2-7} \cline{9-10} 
 & \multicolumn{1}{c|}{TF-IDF} & \multicolumn{1}{c|}{BOW} & CountVectorizer & \multicolumn{1}{c|}{Word2vec} & \multicolumn{1}{c|}{Glove} & FastText &  & \multicolumn{1}{c|}{cTakes} & Metamap \\ \hline
Clinical notes classification & \multicolumn{1}{c|}{$\sqrt{ }$} & \multicolumn{1}{c|}{$\sqrt{ }$} & $\chi$ & \multicolumn{1}{c|}{$\sqrt{ }$} & \multicolumn{1}{c|}{$\chi$} & $\chi$ & $\chi$ & \multicolumn{1}{c|}{$\sqrt{ }$} & $\chi$ \\ \hline
DL assessment for ICD & \multicolumn{1}{c|}{$\sqrt{ }$} & \multicolumn{1}{c|}{$\sqrt{ }$} & $\chi$ & \multicolumn{1}{c|}{$\sqrt{ }$} & \multicolumn{1}{c|}{$\chi$} & $\chi$ & $\chi$ & \multicolumn{1}{c|}{$\chi$} & $\chi$ \\ \hline
Free text classification & \multicolumn{1}{c|}{$\sqrt{ }$} & \multicolumn{1}{c|}{$\chi$} & $\chi$ & \multicolumn{1}{c|}{$\chi$} & \multicolumn{1}{c|}{$\chi$} & $\chi$ & $\chi$ & \multicolumn{1}{c|}{$\chi$} & $\chi$ \\ \hline
Medical text labeling & \multicolumn{1}{c|}{$\sqrt{ }$} & \multicolumn{1}{c|}{$\chi$} & $\chi$ & \multicolumn{1}{c|}{$\chi$} & \multicolumn{1}{c|}{$\chi$} & $\chi$ & $\chi$ & \multicolumn{1}{c|}{$\chi$} & $\chi$ \\ \hline
recognising alcohol consumption & \multicolumn{1}{c|}{$\sqrt{ }$} & \multicolumn{1}{c|}{$\chi$} & $\chi$ & \multicolumn{1}{c|}{$\chi$} & \multicolumn{1}{c|}{$\chi$} & $\chi$ & $\chi$ & \multicolumn{1}{c|}{$\chi$} & $\chi$ \\ \hline
Computerised ICD coding & \multicolumn{1}{c|}{$\sqrt{ }$} & \multicolumn{1}{c|}{$\chi$} & $\chi$ & \multicolumn{1}{c|}{$\chi$} & \multicolumn{1}{c|}{$\chi$} & $\chi$ & $\chi$ & \multicolumn{1}{c|}{$\chi$} & $\chi$ \\ \hline
Indexing biomedical literature & \multicolumn{1}{c|}{$\sqrt{ }$} & \multicolumn{1}{c|}{$\chi$} & $\chi$ & \multicolumn{1}{c|}{$\chi$} & \multicolumn{1}{c|}{$\chi$} & $\chi$ & $\chi$ & \multicolumn{1}{c|}{$\chi$} & $\chi$ \\ \hline
Multi-label classification & \multicolumn{1}{c|}{$\chi$} & \multicolumn{1}{c|}{$\sqrt{ }$} & $\chi$ & \multicolumn{1}{c|}{$\chi$} & \multicolumn{1}{c|}{$\chi$} & $\chi$ & $\chi$ & \multicolumn{1}{c|}{$\chi$} & $\chi$ \\ \hline
Clinical coding & \multicolumn{1}{c|}{$\chi$} & \multicolumn{1}{c|}{$\sqrt{ }$} & $\chi$ & \multicolumn{1}{c|}{$\chi$} & \multicolumn{1}{c|}{$\chi$} & $\chi$ & $\chi$ & \multicolumn{1}{c|}{$\chi$} & $\chi$ \\ \hline
ML-based encoding & \multicolumn{1}{c|}{$\chi$} & \multicolumn{1}{c|}{$\sqrt{ }$} & $\chi$ & \multicolumn{1}{c|}{$\chi$} & \multicolumn{1}{c|}{$\chi$} & $\chi$ & $\chi$ & \multicolumn{1}{c|}{$\chi$} & $\chi$ \\ \hline
Feature identification & \multicolumn{1}{c|}{$\chi$} & \multicolumn{1}{c|}{$\sqrt{ }$} & $\chi$ & \multicolumn{1}{c|}{$\chi$} & \multicolumn{1}{c|}{$\chi$} & $\chi$ & $\chi$ & \multicolumn{1}{c|}{$\chi$} & $\chi$ \\ \hline
Extracting medication & \multicolumn{1}{c|}{$\chi$} & \multicolumn{1}{c|}{$\chi$} & $\chi$ & \multicolumn{1}{c|}{$\sqrt{ }$} & \multicolumn{1}{c|}{$\chi$} & $\chi$ & $\chi$ & \multicolumn{1}{c|}{$\chi$} & $\chi$ \\ \hline
ICD encoding & \multicolumn{1}{c|}{$\chi$} & \multicolumn{1}{c|}{$\chi$} & $\chi$ & \multicolumn{1}{c|}{$\sqrt{ }$} & \multicolumn{1}{c|}{$\chi$} & $\chi$ & $\chi$ & \multicolumn{1}{c|}{$\chi$} & $\chi$ \\ \hline
Clinical coding & \multicolumn{1}{c|}{$\chi$} & \multicolumn{1}{c|}{$\chi$} & $\chi$ & \multicolumn{1}{c|}{$\sqrt{ }$} & \multicolumn{1}{c|}{$\sqrt{ }$} & $\chi$ & $\chi$ & \multicolumn{1}{c|}{$\chi$} & $\chi$ \\ \hline
Note classification & \multicolumn{1}{c|}{$\chi$} & \multicolumn{1}{c|}{$\chi$} & $\chi$ & \multicolumn{1}{c|}{$\chi$} & \multicolumn{1}{c|}{$\sqrt{ }$} & $\chi$ & $\chi$ & \multicolumn{1}{c|}{$\chi$} & $\chi$ \\ \hline
Note embedding & \multicolumn{1}{c|}{ $\chi$} & \multicolumn{1}{c|}{ $\chi$} &  $\chi$ & \multicolumn{1}{c|}{ $\chi$} & \multicolumn{1}{c|}{$\sqrt{ }$} &  $\chi$ &  $\chi$ & \multicolumn{1}{c|}{ $\chi$} &  $\chi$ \\ \hline
Classifying diagnosis & \multicolumn{1}{c|}{ $\chi$} & \multicolumn{1}{c|}{ $\chi$} &  $\chi$ & \multicolumn{1}{c|}{ $\chi$} & \multicolumn{1}{c|}{$\sqrt{ }$} &  $\chi$ &  $\chi$ & \multicolumn{1}{c|}{ $\chi$} &  $\chi$ \\ \hline
Pre-screening for paediatric oncology patients & \multicolumn{1}{c|}{$\chi$} & \multicolumn{1}{c|}{$\chi$} & $\chi$ & \multicolumn{1}{c|}{$\chi$} & \multicolumn{1}{c|}{$\chi$} & $\chi$ & $\chi$ & \multicolumn{1}{c|}{$\sqrt{ }$} & $\chi$ \\ \hline
DL comparison & \multicolumn{1}{c|}{$\chi$} & \multicolumn{1}{c|}{$\chi$} & $\chi$ & \multicolumn{1}{c|}{$\chi$} & \multicolumn{1}{c|}{$\chi$} & $\chi$ & $\chi$ & \multicolumn{1}{c|}{$\sqrt{ }$} & $\chi$  \\ \hline
Feature engineering & \multicolumn{1}{c|}{$\chi$ } & \multicolumn{1}{c|}{$\chi$ } & $\chi$  & \multicolumn{1}{c|}{$\chi$ } & \multicolumn{1}{c|}{$\chi$ } & $\chi$  & $\chi$  & \multicolumn{1}{c|}{$\sqrt{ }$} & $\chi$  \\ \hline
Knowledge extraction & \multicolumn{1}{c|}{$\chi$} & \multicolumn{1}{c|}{$\chi$} & $\chi$ & \multicolumn{1}{c|}{$\chi$} & \multicolumn{1}{c|}{$\chi$} & $\chi$ & $\chi$ & \multicolumn{1}{c|}{$\sqrt{ }$} & $\chi$ \\ \hline
Text mining of cancer & \multicolumn{1}{c|}{ $\chi$} & \multicolumn{1}{c|}{$\chi$ } & $\chi$  & \multicolumn{1}{c|}{$\chi$ } & \multicolumn{1}{c|}{$\chi$ } & $\chi$  & $\chi$  & \multicolumn{1}{c|}{$\sqrt{ }$} & $\chi$  \\ \hline
Free text analysis & \multicolumn{1}{c|}{$\chi$} & \multicolumn{1}{c|}{$\chi$} & $\chi$ & \multicolumn{1}{c|}{$\chi$} & \multicolumn{1}{c|}{$\chi$} & $\chi$ & $\chi$ & \multicolumn{1}{c|}{$\sqrt{ }$} & $\chi$ \\ \hline
Extraction of drugs indications & \multicolumn{1}{c|}{$\chi$ } & \multicolumn{1}{c|}{$\chi$ } & $\chi$  & \multicolumn{1}{c|}{$\chi$ } & \multicolumn{1}{c|}{$\chi$ } & $\chi$  & $\chi$  & \multicolumn{1}{c|}{$\chi$ } & $\sqrt{ }$ \\ \hline
Diagnosis codes to free-text & \multicolumn{1}{c|}{$\chi$ } & \multicolumn{1}{c|}{$\chi$ } & $\chi$  & \multicolumn{1}{c|}{$\chi$ } & \multicolumn{1}{c|}{$\chi$ } & $\chi$  & $\chi$  & \multicolumn{1}{c|}{$\chi$ } & $\sqrt{ }$ \\ \hline
\end{tabular}%
}
\end{table}

\subsection{Automated tools}
\label{CSC}
This sub-section will explain the automated tools currently being utilised in healthcare. Tables \ref{tab:my-table_IT} and \ref{tab:my-table_CA} illustrate automated machine learning integrated tools employed for commercial and open-source applications. Overall, it is evident that several automated ML solutions produced by Google, Amazon, Microsoft, and JADBIO are chargeable and do not require coding. Likewise, technologies that do not require a fee necessitate minimum coding in the local environment and have some limits compared to fee-based solutions. On the other hand, existing AutoML technologies are not commercially available for structured data and focus primarily on well-defined unstructured data.

\vspace{3mm}
\begin{longtable}[c]{|c|c|c|c|c|c|}
\caption{Automated machine learning integrated tools.}
\label{tab:my-table_IT}\\
\hline
Platform & Chargeable & Coding & Environment & Dataset & Domain \\ \hline
\endfirsthead
\multicolumn{6}{c}%
{{\bfseries Table \thetable\ continued from previous page}} \\
\hline
Platform & Chargeable & Coding & Environment & Dataset & Domain \\ \hline
\endhead

Auto   ML\cite{faes2019automated} & $\sqrt{ }$ & $\chi$ & Google   Cloud & Images,   Text, and Tabular & Nonspecific \\ \hline
Create   ML\cite{faes2019automated} &  $\chi$ & $\sqrt{ }$ & Local & Images,   Text, and Tabular & Nonspecific \\ \hline
Amazon   Auto ML\cite{faes2019automated} & $\sqrt{ }$ &  $\chi$ & Cloud & Images,   Text, and Tabular & Nonspecific \\ \hline
Microsoft   Auto ML\cite{faes2019automated} & $\sqrt{ }$ &  $\chi$ & Cloud & Images,   Text, and Tabular & Nonspecific \\ \hline
Auto-Sklearn\cite{feurer2015efficient} &  $\chi$ & $\sqrt{ }$ & Local & Tabular & Nonspecific \\ \hline
Auto-WEKA\cite{kocbek2019using} &  $\chi$ & $\sqrt{ }$ & Local & Tabular & Nonspecific \\ \hline
Auto-Keras\cite{li2017hyperband} &  $\chi$ & $\sqrt{ }$ & Local & Tabular & Nonspecific \\ \hline
TPOT\cite{olson2016tpot} &  $\chi$ & $\sqrt{ }$ & Local & Tabular & Nonspecific \\ \hline
JADBIO\cite{Tsamardinos2020.05.04.075747} & $\sqrt{ }$ &  $\chi$ & Cloud & Tabular & Biomedical   and Multi-omics \\ \hline
AutoPrognosis\cite{alaa2018autoprognosis} &  $\chi$ & $\sqrt{ }$ & Local & Tabular & Biomedical \\ \hline
\end{longtable}

Observing Table \ref{tab:my-table_IT}, numerous platforms have created automated ML-enabled solutions for a variety of activities, with Auto-Sklern \cite{feurer2015efficient} being the most popular because it is integrated into the Sklearn library and is designed to select algorithms and optimise hyperparameters~\cite{rana2011adaptive}. This approach also uses Bayesian optimisation techniques and meta-learning to perform its tasks. Another open-source platform developed by the University of British Columbia is Auto-Weka \cite{kocbek2019using}, sometimes known as the Automated Waikato Environment for Knowledge Analysis. Auto-Weka uses Bayesian optimisation for hyperparameter optimisation. The AutoML platform, which utilises statistical algorithms, can only analyse structured data, such as stock market prices, student grades, hotel occupancy, etc.

\vspace{5mm}
\begin{table}[H]
\centering
\caption{Commercially available and as open source automated machine learning tools used.}
\label{tab:my-table_CA}
\begin{tabular}{|c|c|p{6.3cm}|ccc|}
\hline
Dataset Format & Category & Feature & \multicolumn{3}{c|}{Platform} \\ \hline
\multirow{7}{*}{Unstructured} & \multirow{4}{*}{Audio} &  & \multicolumn{1}{c|}{Health Associated} & \multicolumn{1}{c|}{Open Source} & Commercial \\ \cline{3-6} 
 &  & Hearing Aid \cite{bhat2020automated} & \multicolumn{1}{c|}{$\chi$} & \multicolumn{1}{c|}{$\chi$} & $\sqrt{ }$ \\ \cline{3-6} 
 &  & Lung cancer\cite{borkowski2019google} & \multicolumn{1}{c|}{$\chi$} & \multicolumn{1}{c|}{$\chi$} & $\sqrt{ }$ \\ \cline{3-6} 
 & \multirow{3}{*}{Images} & Generic \cite{weng2019unet}\cite{faes2019automated} & \multicolumn{1}{c|}{$\sqrt{ }$} & \multicolumn{1}{c|}{$\chi$} & $\chi$ \\ \cline{3-6} 
 &  & Liver Injury \cite{puri2020automated} & \multicolumn{1}{c|}{$\chi$} & \multicolumn{1}{c|}{$\chi$} & $\sqrt{ }$ \\ \cline{3-6} 
 &  & Ophthalmic syndrome \cite{kim2021classification} & \multicolumn{1}{c|}{$\chi$} & \multicolumn{1}{c|}{$\chi$} & $\sqrt{ }$ \\ \hline
\multirow{8}{*}{Structured} & \multirow{8}{*}{Tabular} & Alzheimer’s disease\cite{karaglani2020accurate} & \multicolumn{1}{c|}{$\sqrt{ }$} & \multicolumn{1}{c|}{$\chi$} & $\chi$ \\ \cline{3-6} 
 &  & BioSignature \cite{tsamardinos2020just} & \multicolumn{1}{c|}{$\sqrt{ }$} & \multicolumn{1}{c|}{$\sqrt{ }$} & $\chi$ \\ \cline{3-6} 
 &  & Brain Age \cite{dafflon2020automated} & \multicolumn{1}{c|}{$\chi$} & \multicolumn{1}{c|}{$\sqrt{ }$} & $\chi$ \\ \cline{3-6} 
 &  & Brain Tumor \cite{su2020automated} & \multicolumn{1}{c|}{$\chi$} & \multicolumn{1}{c|}{$\sqrt{ }$} & $\chi$ \\ \cline{3-6} 
 &  & Cardiovascular disease prognosis \cite{alaa2018autoprognosis} & \multicolumn{1}{c|}{$\sqrt{ }$} & \multicolumn{1}{c|}{$\sqrt{ }$} & $\chi$ \\ \cline{3-6} 
 &  & Diabetes \cite{kocbek2019using} & \multicolumn{1}{c|}{$\chi$} & \multicolumn{1}{c|}{$\sqrt{ }$} & $\chi$ \\ \cline{3-6} 
 &  & Generic \cite{luo2017automating} & \multicolumn{1}{c|}{$\sqrt{ }$} & \multicolumn{1}{c|}{$\sqrt{ }$} & $\chi$ \\ \cline{3-6} 
 &  & Metabolic \cite{ooms2020self} & \multicolumn{1}{c|}{$\chi$} & \multicolumn{1}{c|}{$\sqrt{ }$} & $\chi$ \\ \hline
\end{tabular}
\end{table}

It can also be seen that the majority of unstructured data is processed with commercial AutoML solutions, while structured data is often processed with open-source tools and clinical AutoML platforms. Due to its process-readiness, structured information is easier to manipulate, which is a matter of great concern in this regard. On the other hand, business organisations that benefit from AutoML platforms have created more dynamic text and image processing capabilities. Some examples are Amazon's Recognizer, Apple's CreateML, Microsoft's AutoML, and Google's AutoML. These industries have invested heavily in the implementation of ML-enabled automated platforms. As a result, the majority of their products required no coding, and their tools became relatively user-friendly.

\section{Research Viewpoint}
\label{research_view}

This section discusses current research trends, core challenges in medical NLP, research gaps, and potential future directions.

\subsection{Trend of Current Clinical NLP Research}
\label{research_trend}

\begin{figure}[H]
    \centering
    \includegraphics[width=0.9\textwidth]{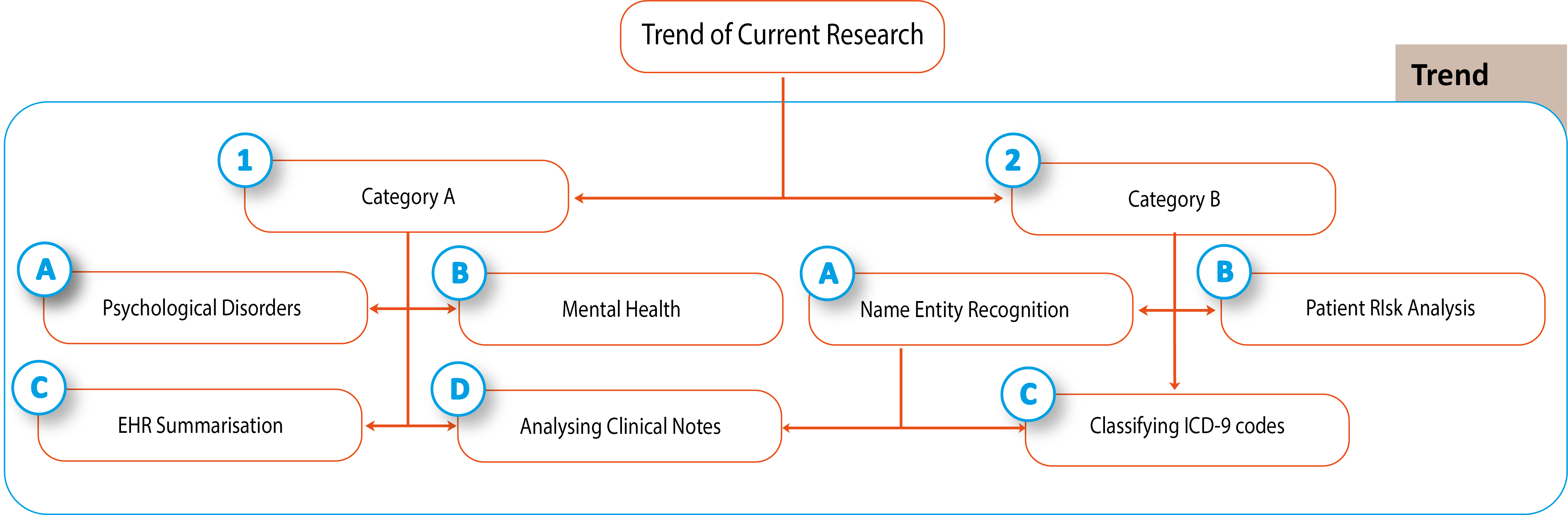}
    \captionsetup{justification=centering}
    \caption{Trend of current medical NLP research.}
    
    \label{fig:research_trend}
\end{figure}

Current trends in medical NLP research are illustrated in Figure~\ref{fig:research_trend}. Analysis of clinical notes from five studies \cite{ayre2021developing} \cite{tsui2021natural} \cite{feller2018using} \cite{feller2018using} \cite{deng2021natural} revealed a substantial effort was put into patient risk assessment. We found several papers that explored ICD-9 code classification; however, most of the articles did not emphasise machine learning or deep learning techniques. As a result, we discarded them because they did not meet our criteria for literature selection. We only found two publications \cite{li2018automated} \cite{nigam2016applying} that emphasised ICD-9 code classification based on machine learning and deep learning.

In addition, the scientific community is currently paying considerable attention to clinical Named Entity Recognition (NER) and medical text summarisation. Five studies \cite{portet2009automatic} \cite{moradi2019small} \cite{liu2018unsupervised} \cite{mcinerney2020query} \cite{molla2020query} created the medical text summarisation model, whereas two studies  \cite{med7} \cite{gligic2020named} offered the medical NER model. The most prevalent physiological illnesses in recent years were dementia and geriatric mental health. There were five studies on dementia \cite{miled2020predicting} \cite{tsang2020modeling} \cite{shao2019detection} \cite{ford2021automated} \cite{wang2019development} and three on geriatric mental health \cite{anzaldi2017comparing} \cite{kharrazi2018value} \cite{chen2019identifying}. Many studies have come up with solutions to alleviate such disorders, yet, there is a significant research opportunity in this field.

\subsection{Core Challenges in Clinical NLP}

In clinical NLP, the core challenge is information overload, which poses a substantial problem in accessing a particular, significant piece of information from vast datasets. In addition, semantic and context understanding are essential and challenging for summarisation systems with a quality deficiency and issues related to usability. Also another significant problem is the wide variety of text formats that an NLP program has to deal with to answer queries from several sources. The following subsections provide a detailed description of the challenges in clinical natural language processing.

\begin{figure}[H]
    \centering
    \captionsetup{justification=centering}
    \includegraphics[width=0.9\textwidth]{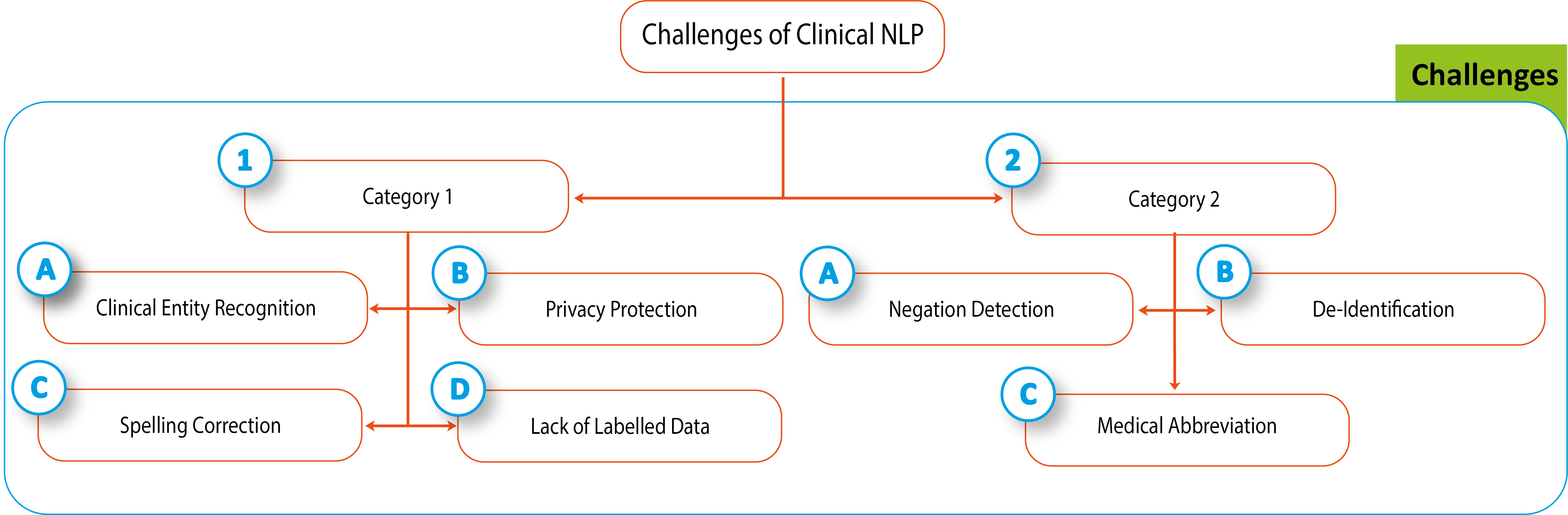}
    \caption{Potential challenges of clinical natural language processing.}
    
    \label{fig:core_challenge}
\end{figure}

\label{med_challenges}

\subsubsection{Medical abbreviations}

Sometimes abbreviations are misread, misinterpreted or misunderstood. Their use increases the amount of time needed to train health care professionals, wastes time determining their cost, often delays patient care, and occasionally results in patient harm. As shown in Figure ~\ref{fig:core_challenge}, many studies have combined medical abbreviations, confusing terms, or data from different diseases when analysing clinical descriptions as a strategy to overcome NLP challenges. Researchers have used these strategies to address frequent clinical phrases used to describe patient care: time of admission, time of initial examination, time of hospitalisation, time to discharge planning, and throughout the coding and billing process \cite{sinha2011use}. Similar challenges are presented by medical abbreviations and acronyms, such as when prescribing medications. Physicians sometimes use Latin-derived abbreviations to specify the frequency of drug administration, such as "BD" (bis die), meaning twice daily. Computers had trouble correctly recognising patterns when identifying such complex abbreviations\cite{jaber2022disambiguating}. However, no study has effectively filtered medical acronyms by removing stop words.

\subsubsection{Spelling Correction and Negation Detection}

Terms in medical summary documents are misspelt for two reasons: clerical errors and Optical Character Recognition (OCR) errors. The Levenshtein distance is a straightforward method for replacing all misspelt words with dictionary words. While many dictionary words can have the same edit distance, replacing words based solely on Levenshtein distance \cite{black2008levenshtein} does not result in higher accuracy. Correcting misspelt words with those the language model suggests significantly improves accuracy. Since medical texts use a distinct language than other texts, language models developed using general English do not work well in this case. 

Shannon's Noisy Channel Model \cite{lai2015automated} is the most effective development in NLP in terms of recent advances in medical spelling correction. This technique uses an extensive dictionary built from numerous sources. This model predicates named entity recognition so that misspellings are not incorrectly corrected. This spell checker was applied to three distinct forms of free-text data: clinical notes, allergy entries, and pharmaceutical orders. The efficacy of this model is that it is capable of high-performance spelling correction in various clinical narratives. To the best of our knowledge, no suitable tools for medical spelling checking have yet been widely produced; therefore, it is essential to build further tools, such as Shannon's noisy channel model \cite{lai2015automated}, to overcome the highlighted issues.

On the other hand, a linguistic phenomenon known as negation causes sentences to have their meanings reversed. The negation term determines if a finding in the clinical narrative has to be annotated as a finding or should be excluded. For example, Kundeti et al. \cite{kundeti2016clinical} demonstrated that the significance of the findings is altered by the use of qualifiers and negation terms. For instance, a cyst is a finding in the statement "cyst detected in the lungs," but it is no longer a finding in the sentence "No cyst identified in the lungs." Another example NegEx developed by Mehrabi et al. \cite{mehrabi2015deepen}. NegEx is an algorithm for negation detection that has proven effective in clinical NLP. NegEx fails to accurately determine the negation status of concepts in complex phrases because it disregards the contextual relationship between words inside a sentence. 

\subsubsection{Lack of Medical Data}

Data shortages have become a significant obstacle for medical NLP research~\cite{latif2022multitask}. Adopting supervised ML models successfully solves a variety of healthcare challenges, and sufficient training data is a precondition for deploying supervised machine learning algorithms. However, many health systems are hesitant to share confidential patient data due to ethical, privacy, and liability concerns.

Most of the study data in studies we reviewed were acquired directly from hospitals rather than from more convenient and accessible online repositories such as Kaggle or the UCI Machine Learning Repository; hence, data scarcity is now a hurdle for the scientific community, particularly for clinical NLP research. Therefore, modern techniques such as transformer-based models and cutting-edge deep learning algorithms are not commonly utilised in this field. This leads us to the following question:  how much data is required to perform research in medical NLP? The minimum amount of data needed for an AI study cannot be established with any degree of accuracy. It goes without saying that the nature of a project significantly impacts how much data is needed. Text, images, and videos, for instance, typically require considerable data. To generate an accurate estimate, however, additional criteria, such as the number of anticipated categories and model performance, must be addressed.

\subsubsection{Sensitivity of Medical Data and Privacy}

Clinical notes contain detailed information about patient-physician interactions. During these exchanges, patients reveal their health difficulties, eating habits and potentially stigmatising disorders. The Health Insurance Portability and Accountability Act (HIPAA) privacy law guarantees the privacy of personal medical information in the United States. In addition, the European Union's General Data Protection Regulation (GDPR) establishes regulations for using health data for scientific purposes. These legislative moves have immediate implications for NLP research, with informed consent from individuals and sanitisation of sensitive data categories being paramount.
The GDPR outlines broad principles regarding the processing of confidential data, including that the processing must be fair, transparent, and lawful (i.e., with consent), carried out for specific and legitimate purposes, and the data should be retained for no longer than is required. This is known as data minimisation, which includes sanitisation. "Special categories of personal data are primarily employed in the scientific analysis." The processing of private information is only permissible with the subject's express consent or after the person has made the information public. Generally, "scientific usage" refers to basic, applied, privately funded research and technological innovation.

Methods of sanitisation are frequently regarded as the bare minimum for protecting the privacy of individuals when collecting data. The goal is to utilise a technology that generates entirely new copies of the dataset that appear real for data analysis while protecting the privacy of the individuals in the dataset to a certain extent, depending on the technique. The sanitisation approach has been criticised for numerous reasons, even though it is a vital step in protecting patient privacy. Initially, both the data's value and integrity are compromised. Second, while sanitisation promotes data access and sharing, achieving this is not always adequate. This is primarily due to the possibility that the deductive discovery could result in the re-identification of the original sensitive data.

\subsection{Future Research Directions}
This sub-section highlights gaps and provides future research directions for various aspects of natural language processing in electronic health records. 

\subsubsection{Model Assessment and Point of View of Adopted Models}

The existing literature did not validate their proposed models using model validation indicators, such as K-Fold Cross Validation, as the first point in this regard. A model may contain both generalisation and overfitting errors. When a model is overfitted, it performs exceptionally well on the training data but fails when presented with new data. Generalisation is the term used to describe a model's performance on new data. Therefore, evaluating ML models before applying them to an algorithm is essential. Note that K– Fold Cross Validation is an effective method for mitigating these underlying problems \cite{wong2019reliable}. The K– Fold Cross Validation method is used to develop an ML model on multiple subsets of the same dataset, resulting in different prediction accuracy for each subgroup. This approach allows us to assess how much the accuracy performance of a model differs for distinct data and the average accuracy of other data. Cross Validation has distinctive characteristics, such as the fact that different folds have various efficacy, so it is possible to estimate how well a proposed model will perform overall, and overfitting can be eliminated using this technique \cite{santos2018cross}. Also, regularization strategies like as dropout, L1/L2, capacity reduction, and early stopping are primarily used to combat overfitting.

In contrast, TF-IDF and BOW approaches were used in many of the reviewed research articles to extract features from the free text in EHRs, indicating that relatively little research has utilised cutting-edge word embedding techniques. The word2vec method was utilised in a few articles, but its prevalence was negligible compared to traditional feature extraction methods. Given conventional feature extraction methods, acquiring semantic information is challenging; however, this difficulty can be mitigated by introducing other advanced word embedding methods such as FastText, Glove, and BERT.

In recent years, this has been a top priority for clinical note analyses, including the identification of goal-of-care documentation in EHRs and suicide prediction from large-scale clinical records. A number of deficiencies must be addressed in order to improve the adopted solutions, despite the satisfactory performance achieved in these contexts. Concerning the identification of goal-of-care discussions in EHRs, it is evident that algorithms with extremely high positive predictive values may not achieve sufficient positive predictive values. Future research should consider using a word- or phrase-level annotation of training data (such as identifying specific sections of a note containing documentation of goal-care), developing ontologies to interpret goal-care discussions, and investigating cutting-edge techniques to eliminate training data biases that differ from real-world data.

Regarding suicidal prediction, a multitude of obstacles has hindered understanding, forecasting, and preventing suicidal behaviour. First, there is a lack of knowledge regarding the actual suicide predictors \cite{walsh2018predicting}. Numerous risk factors have been identified, including mental illness, youth, and a history of suicide behaviour. However, these characteristics have limited predictive accuracy for suicidal conduct and account for a minor percentage of the variance. Second, there is no recognised model for identifying the relationship between risk factors and suicide conduct. Apart from these, rarity of suicide - rare events is harder to predict. It's important to make a distinction between suicide and suicide attempts, which re more common but might not share the same predictors. This is an issue for researchers and clinicians because there is no technique that doctors can use to combine data when determining if a patient is likely to attempt suicide in the near future. Medical practitioners must rely on intuition, which is no more reliable than random chance in predicting suicide behaviour. On the basis of the above discussion, it is necessary to obtain longitudinal data from large data samples that can be used to create and test novel models of suicide risk. However, if suicide researchers examine the notion, it will indicate a possible new direction for future research and clinical treatment.

\subsubsection{Perspectives on ML integration and AutoML}
Various businesses hire data scientists for data processing and decision-making \cite{smaldone2022employability}. Data scientists sometimes lack interdisciplinary experience, particularly in clinical natural language processing \cite{sarker2021defining}. It is uncommon to find experts with such insight.  

Although AI-integrated tools are widely accepted, several flaws, such as business challenges and security, have been identified. Security is one of the most frequently discussed areas of research in AutoML. Regarding the security of AutoML, businesses are investigating various technological solutions, such as automatic machine learning for privacy protection, automated multi-party machine learning, automated federation, etc. However, the current implementation does not support laws, regulations, and industry standards. Specifically, federated learning, also known as collaborative learning, is an ML technique that uses several distributed edge devices or servers that store local data samples to train an algorithm without transferring the data samples. Multi-Party Computing (MPC) is a cryptographic technique that allows multiple parties to perform computations using combined data without revealing their inputs. Currently, these methods address various privacy and security concerns \cite{li2020privacy}. Therefore, organisations must encourage developing and enhancing standards for federated learning and secure multi-party computing.

In addition to these issues, AutoML typically encounters problems with data and model applications. For instance, insufficient high-quality labelled data and data inconsistencies will hurt offline data analysis. Automating the processing of unstructured and semi-structured data by machine learning is necessary but technically challenging. The current optimisation objectives for the AutoML system are predefined. Multiple purposes, such as differentiating between decision-making and cost, frequently present a challenge. This multi-objective investigation has limited analysis options before yielding effective results. The actual business may have specialised data processing requirements for the existing machine learning process. In the current Black Box AutoML solution, such conditions are poorly handled. Consequently, it may be necessary to adopt a new solution capable of mitigating the shortcomings of the current Black Box system.

On the contrary, many studies indicate that researchers frequently concentrate on developing a model rather than deploying it in a production environment. Implementing a simulation and deploying it in production are two distinct processes, as a model may perform well during the simulation phase but contain numerous errors when analysing real-time data. Therefore, model deployment is essential for determining the efficacy of proposed models. If an End-to-end model can be developed, it will significantly contribute to clinical settings and provide an effective tool for more efficiently analysing complex patient data.

\subsubsection{Medical Data Imbalance and Data Shortage}

In Table \ref{tab:data_processing}, we demonstrate that researchers did not specify which text preprocessing pipeline settings were selected to handle unstructured EHRs. This is an essential deficiency in reporting this type of research, as cleaning medical text differs significantly from cleaning other data. For example, text normalisation or stop word removal methods are available that perform correctly, but due to various abbreviations and terms in medical free text, applying current stop word or normalisation approaches is somewhat challenging and limits the ability to obtain accurate results. Likewise, when working with any text data, an imbalanced dataset creates several issues \cite{lu2022comparative}. Typically, "imbalanced data" refers to a categorisation issue in which the classes are not represented equally. Before applying the data to the ML system, it is recommended to address the imbalance issue in clinical free text in EHRs. The problem of data imbalance can be alleviated by utilising techniques such as the Synthetic Minority Oversampling Technique (SMOTE). This can increase the number of cases from the dataset and reduce the medical data imbalance problem by combining oversampling and undersampling \cite{wang2021cab}. In addition, Class disparity can be addressed through cost-sensitive training and sampling technique. However, a significant deficiency in the reviewed literature was that the solutions to address these issues were not described. 

On the other hand, the scarcity of medical data is one of the current obstacles. Utilising synthetic data is a possible solution \cite{guan2018generation, latif2021survey,latif2020augmenting}. It may provide a safer method of development for clinical data. Synthetic data is frequently employed when there is insufficient actual data or not enough to identify specific patterns. Both training and testing datasets utilise it in the same manner. Transfer learning techniques can be used as a substitute when there is an absence of training data for the target domain, and there are few or no exact matches between the source and target domains. Lastly, the Naive Bayes algorithm, one of the simplest classifiers, should be more widely recognised for its utility when dealing with clinical data, as it learns surprisingly well from relatively small data sets.

\subsection{Limitations of the study}

A limitation of our study is that we did not consider grey literature, which consists of academic papers such as theses and essays, research and committee reports, official reports, conference articles, and ongoing research \cite{paez2017gray}. Compared to scientific research, grey literature publications might be a more detailed source of information as they can be longer and include more information because a typical structure does not constrain them. Due to the heterogeneity of the papers, no meta-analyses were included in this review. Another major weakness of this study is the failure to assess publication bias which can occur for several reasons. Some researchers may decide not to publish their findings if they discover that the data sets do not support their hypothesis. In this case, they prefer to present study reports that support their incorrect hypothesis. When publication bias becomes widespread, favourable results are overrepresented in the scientific literature, impairing our comprehension of any systematic investigation. Comparing the findings of published and unpublished research on the same subject is an efficient method for identifying publication bias. Comparing results can reveal if there is a positive result bias across studies. As understanding current clinical NLP challenges and the techniques used to analyse EHRs was our primary objective, we did not assess publication bias at this stage.

\section{Conclusion}
\label{conclusion}

This study finds that recent advancements in Machine Learning and Deep Learning models can facilitate health informatics tasks on Electronic Health Records (EHR). We have concentrated on conducting a thorough analysis of natural language processing in electronic health records. We reviewed recent research on the following EHRs-NLP tasks: patient risk analysis/prediction, state-of-the-art architectures for analysing EHRs, medical text summarisation, and other NLP applications such as clinical named entity recognition, blockchain-based EHRs, mental health research, goals of care conversations, clinical chart review, negation identification, and medical language translation. In addition, we provide a list of automated ML-enabled tools used by the healthcare industry and medical experts to support EHR-NLP research. The highlight of our findings are as follows: 

\begin{enumerate}
\item Physiological disorders, such as dementia and geriatric mental health, have been identified as promising research areas and are the subject of ongoing research in which various models and methods for extracting features suited to these tasks are being explored. 

\item The literature review performed in this work shows that SVM, boosting techniques, LR, LSTM, RNN, and CNNs are appropriate for analysing unstructured free text data for downstream EHRs applications.

\item We find that while deep learning algorithms have achieved great success in the NLP sector, their application in the biological realm remains difficult. In contrast to classic ML models, which are frequently used for health records, DL models present a number of disadvantages relating to data availability, the difficulty of domain-specific textual data, and interpretability. Notable is that DL-based algorithms require a large amount of data to outperform other methods, as well as expensive GPUs and hundreds of workstations.

\item Cutting-edge NLP methods, such as transformer-based models for free text analysis, are yet to be used extensively, and conventional methods are currently preferred. Therefore, we wonder if transformer-based techniques will become the de facto standard for clinical NLP. 

\end{enumerate}

\section*{Author Contributions}

Elias Hossain proposed the study, and Niall Higgins subsequently outlined the inclusion and exclusion criteria for downloading and organising the papers. Then, Elias Hossain, under the instructions of Rajib Rana, conducted a literature search and completed a full-text review, finding gaps, challenges, citations, data analysis, illustrations, and future directions. The study was reviewed and edited by Rajib Rana, Niall Higgins, Jeffrey Soar, Prabal Datta Barua, Anthony R. Pisani, and Kathryn Turner, and the corresponding author was notified of additional corrections and clarifications.

\section*{Funding Statement}

This research received no specific grant from any funding agency in the public, commercial, or not-for-profit sectors.

\section*{Declaration of Competing Interest}
The authors of this study declare that they
have no conflict of interest.

\begin{appendix}

\vspace{5mm}

\begin{table}[H]
\caption{Commonly utilised technical terms used in this systematic review}
\label{tab:tech_terms}
\resizebox{\textwidth}{!}{%
\begin{tabular}{|c|l|}
\hline
Technical Terms         & \multicolumn{1}{c|}{Definition}                                                                          \\ \hline
ROC-AUC &
  \begin{tabular}[c]{@{}l@{}}Receiver Operating Characteristics-Area Under The Curve (ROC-AUC).\\ These curve plots the true positive ratio against the false positive rate at \\ different threshold values.\end{tabular} \\ \hline
K– Fold Cross Validation & ML models are created by creating multiple subsets of the same dataset using K– Fold Cross Validation.    \\ \hline
Traditional ML &
  \begin{tabular}[c]{@{}l@{}}Traditional ML models can utilised used to resolve classification, regression, clustering, dimension \\ reduction problems. For examples: linear regression, logistic regression, naive bayes.\end{tabular} \\ \hline
AutoML &
  \begin{tabular}[c]{@{}l@{}}Automated machine learning-enabled tools are used to complete a variety of tasks, \\ especially from clinical tasks to other classification and prediction tasks.\end{tabular} \\ \hline
Feature Extraction      & It is used for transforming raw data into numerical features                                             \\ \hline
Word Embedding          & Word embedding is used for the representation of words for text analysis                                 \\ \hline
TF-IDF, Word2vec, Glove &
  \begin{tabular}[c]{@{}l@{}}Term Frequency-Inverse Document Frequency (TF-IDF).\\ Global Vectors for word representation (Glove). \\ These are used to convert text data to numeric form to to apply the ML algorithm.\end{tabular} \\ \hline
FastText &
  \begin{tabular}[c]{@{}l@{}}Fasttext is an open source, free of charge, lightweight library that lets users learn text representation \\ and text classification.\end{tabular} \\ \hline
BOW                     & Bag of Words (BOW) is a classical word representation technique.                                         \\ \hline
Free Text               & Free text of electronic medical notes is considered a rich source for healthcare operations and research \\ \hline
Overfitting and Underfitting &
  \begin{tabular}[c]{@{}l@{}}Overfitting indicates that a model performs satisfactorily in training data, but performs poorly\\ in new data. However, underfitting works poorly on both datasets.\end{tabular} \\ \hline
Bayesian Optimisation   & Bayesian optimisation methods are effective because they choose hyper parameters in a known manner.      \\ \hline
Stop Words &
  \begin{tabular}[c]{@{}l@{}}In the case of classifying text documents, some terms do not contain the  actual meaning \\ to be used in the classification model. For example: \{"a", "however moreover", "is the", "afterwards","again", etc. .\}\end{tabular} \\ \hline
\end{tabular}%
}
\end{table}

\begin{table}[H]
\centering
\caption{List of the abbreviations used in this manuscript }

\label{tab:my-table}
\begin{tabular}{|l|l|}
\hline
\multicolumn{1}{|c|}{Abbreviations} & \multicolumn{1}{c|}{Meaning}                                       \\ \hline
ROC-AUC                             & Receiver Operating Characteristics-Area Under The Curve            \\ \hline
ML                                  & Machine Learning                                                   \\ \hline
DL                                  & Deep Learning                                                      \\ \hline
NLP                                 & Natural Language Processing                                        \\ \hline
TF-IDF                              & Term Frequency-Inverse Document Frequency                          \\ \hline
Glove                               & Global Vectors for word representation                             \\ \hline
BOW                                 & Bag of Words                                                       \\ \hline
GDPR                                & General Data Protection Regulation                                 \\ \hline
HIPAA                               & The Health Insurance Portability and Accountability Act            \\ \hline
BERT                                & Bidirectional Encoder Representations from Transformers            \\ \hline
CNN                                 & Convolutional Neural Network                                       \\ \hline
TL                                  & Transfer Learning                                                  \\ \hline
ResNet                              & Residual Neural Network                                            \\ \hline
LSTM                                & Long Short-Term Memory                                             \\ \hline
BI-LSTM                             & Bidirectional Long ShortTerm Memory                                \\ \hline
RNN                                 & Recurrent Neural Network                                           \\ \hline
GRU                                 & Gated Recurrent Units                                              \\ \hline
RL                                  & Representation Learning                                            \\ \hline
LR                                  & Logistic Regression                                                \\ \hline
SVM                                 & Support Vector Machine                                             \\ \hline
XGBoost                             & eXtreme Gradient Boosting                                          \\ \hline
RF                                  & Random Forest                                                      \\ \hline
LR                                  & Linear Regression                                                  \\ \hline
NB                                  & Naïve Bayes                                                        \\ \hline
GB                                  & Gradient Boosting                                                  \\ \hline
DT                                  & Decision Tree                                                      \\ \hline
ICD-9                               & The International Classification of Diseases, Ninth Revision       \\ \hline
ICD-10                              & International Classification of Diseases, 10th Revision            \\ \hline
PRISMA                              & Preferred Reporting Items for Systematic Reviews and Meta-Analyses \\ \hline
ICU                                 & Intensive Care Unit                                                \\ \hline
ABLSTM                              & Gated Attention incorporated Bi-Directional Long Short-Term Memory \\ \hline
FCNN                                & Fully Connected Neural Network                                     \\ \hline
MIMIC                               & Medical Information Mart for Intensive Care                        \\ \hline
D2V                                 & Document to Vector                                                 \\ \hline
CUIs                                & Concept Unique Identifiers                                         \\ \hline
CAM                                 & Confusion Assessment Method                                        \\ \hline
LDA                                 & Latent Dirichlet Allocation                                        \\ \hline
HIV                                 & Human Immunodeficiency Virus                                       \\ \hline
LASSO                               & Least Absolute Shrinkage and Selection                             \\ \hline
SMI                                 & Serious Mental Illness                                             \\ \hline
EHR                                 & Electronic Health Record                                           \\ \hline
XML                                 & Extensible Markup Language                                         \\ \hline
CRIS                                & Clinical Record Interactive Search                                 \\ \hline
AMIA                                & American Medical Informatics Association                           \\ \hline
\end{tabular}
\end{table}

\end{appendix}

\bibliographystyle{IEEEtran}
\bibliography{ref}

\end{document}